\documentclass[10pt,twocolumn,letterpaper]{article}

\usepackage[pagenumbers]{cvpr} 

\usepackage{graphicx}
\usepackage{amsmath}
\usepackage{amssymb}
\usepackage{booktabs}
\usepackage{flushend}
\usepackage{cuted}
\usepackage{soul}
\usepackage{bbold}
\usepackage{mwe}
\usepackage[table,dvipsnames]{xcolor}
\usepackage{enumitem}

\definecolor{internationalkleinblue}{rgb}{0.0, 0.18, 0.65}
\usepackage[pagebackref,
            breaklinks,
            colorlinks,
            citecolor=internationalkleinblue]{hyperref}

\usepackage{algorithm2e}

\makeatletter
\@namedef{ver@everyshi.sty}{}
\makeatother
\usepackage{tikz}
\usepackage{pgfplots}\pgfplotsset{compat=1.9}

\usepackage[capitalize]{cleveref}
\crefname{section}{Sec.}{Secs.}
\Crefname{section}{Section}{Sections}
\Crefname{table}{Table}{Tables}
\crefname{table}{Tab.}{Tabs.}

\DeclareMathOperator*{\argmin}{argmin}

\def \ours{\texttt{FOUND}\xspace}  
\def \conv{$conv1\times1$\xspace}
\def \cls{\texttt{CLS}\xspace}  
\def \bs{\zeta}

\def\cvprPaperID{} 
\def\confName{CVPR}
\def\confYear{2023}

\begin{document}

\title{Unsupervised Object Localization:\\Observing the Background to Discover Objects}
\date{}

\author{Oriane Siméoni
\textsuperscript{1}, Chloé Sekkat\textsuperscript{1}, Gilles Puy\textsuperscript{1}, Antonin Vobecky\textsuperscript{1,2}, \'Eloi Zablocki\textsuperscript{1}, Patrick Pérez\textsuperscript{1}\\
\textsuperscript{1}valeo.ai, Paris, France\\
\textsuperscript{2}Czech Institute of Informatics, Robotics and Cybernetics, CTU, Prague, Czech Republic \\
}

\maketitle

\begin{strip}
\centering
    \includegraphics[height=2.93cm]{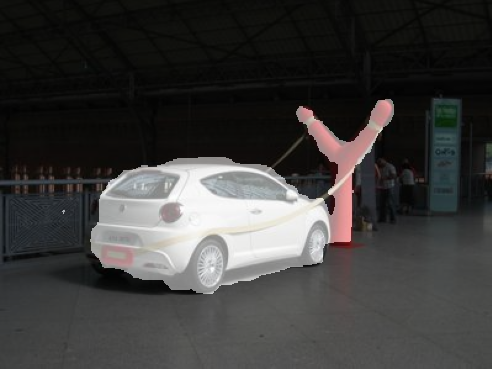}\includegraphics[height=2.93cm]{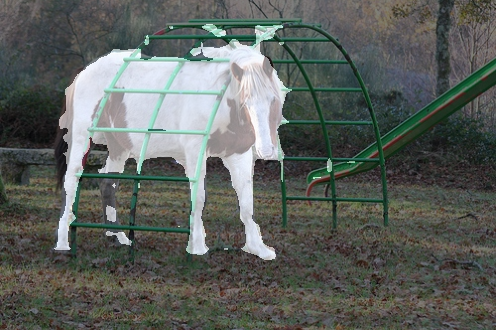}\includegraphics[height=2.93cm]{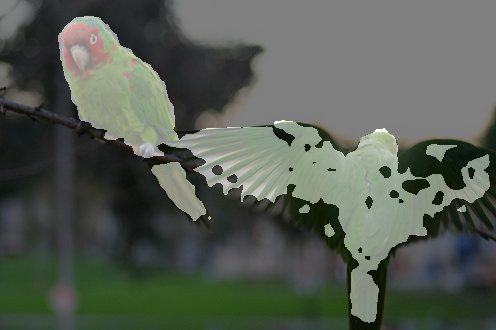}\includegraphics[height=2.93cm]{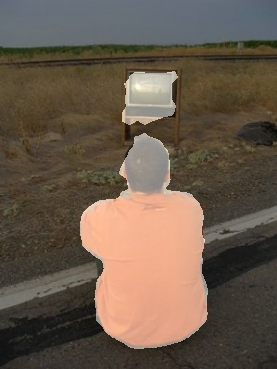}\includegraphics[height=2.93cm]{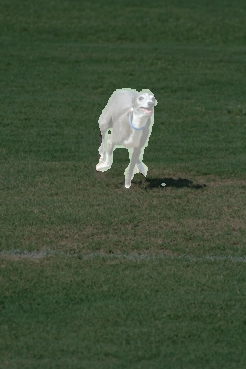}
    \vspace{-4pt}
    \includegraphics[height=3.145cm]{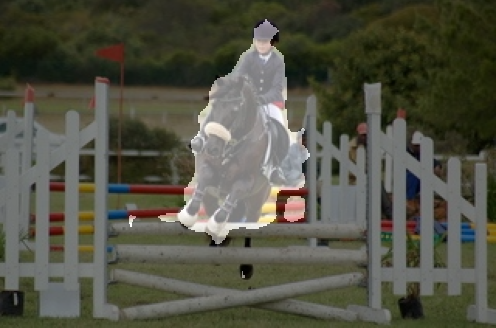}\includegraphics[height=3.145cm]{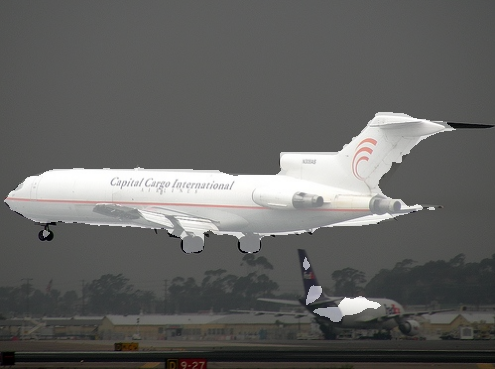}\includegraphics[height=3.145cm]{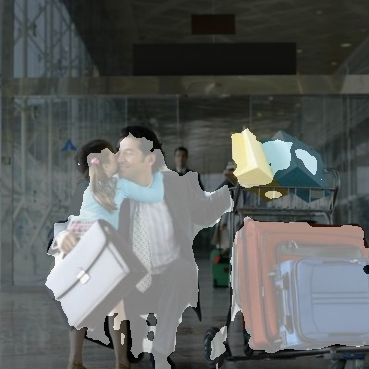}\includegraphics[height=3.145cm]{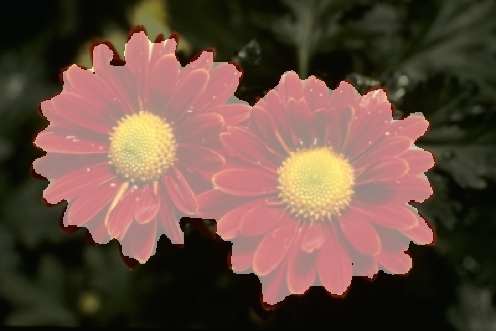}
\captionof{figure}{
\textbf{Examples of object localization results obtained with our method \ours on images from diverse datasets}. We propose a simple framework in which we train a single \emph{\conv} layer, and achieve state-of-the-art results in unsupervised object discovery and saliency detection. We train for only 2 epochs over the 10k dataset DUTS-TR \cite{wang2017} and inference runs at $80$ FPS. Note that the results presented here are without post-processing refinement.}
\label{fig:first}
\end{strip}

\begin{abstract}
Recent advances in self-supervised visual representation learning have paved the way for unsupervised methods tackling 
tasks such as object discovery and instance segmentation.
However, discovering objects in an image with no supervision is a very hard task; what are the desired objects, when to separate them into parts, how many are there, and of what classes? The answers to these questions depend
on the tasks and datasets of evaluation.
In this work, we take a different approach and propose to \textbf{look for the background instead}. This way, the salient objects emerge as a by-product without any strong assumption on what an object should be.
We propose \ours, a simple model made of
a single \textbf{\conv} initialized with coarse background masks extracted from self-supervised patch-based representations.
After fast training and refining these seed masks, the model reaches state-of-the-art results on unsupervised saliency detection and object discovery benchmarks.
Moreover, we show that our approach yields good results in the unsupervised semantic segmentation retrieval task. The code to reproduce our results is available at \url{https://github.com/valeoai/FOUND}.
\end{abstract}

\vspace{-10pt}


\section{Introduction}
\label{sec:intro}

The task of object localization --- either performed by detecting \cite{ren2015fasterrcnn,carion2020detr} or segmenting \cite{chen2018deeplab} objects --- is required in many safety-critical systems such as self-driving cars.  
Today's best methods train large deep models \cite{carion2020detr,chen2018deeplab} on large sets of labeled data \cite{coco2014,pascal-voc-2007}. 
To mitigate such needs in annotation, it is possible to use strategies such as semi-supervised \cite{cheng2021semi,Chen_2021_CVPR}, weakly-supervised \cite{Wu_2021_CVPR,Chen_2021_CVPR} and active learning \cite{Yuan_2021_CVPR,Parvaneh_2022_CVPR, BiB_eccv22}.

In this work, we consider the \textit{unsupervised object localization} task, which consists in discovering objects in an image with no human-made annotation. This task has recently received a lot of attention \cite{seitzer2022dinosaur, shin2022selfmask, Sauvalle2022UnsupervisedMS} as it is a solution to detect objects in a scene with no prior about what they should look like or which category they should belong to. 
Early works exploit hand-crafted features \cite{yan2013hs, zhu2014wctr, uijlings2013selectivesearch, zitnick2014edgebox} and inter-image information \cite{vo2020unsup_multi_object_discovery, vo2021large_scale_unsup_object_discovery} but hardly scale to large datasets.
Recent works leverage strong \emph{self-supervised} \cite{gidaris2021obow, caron2021dino, he2022mae} features learned using pretext tasks:
\cite{simeoni2021lost, wang2022tokencut, melas2022deepsectralmethod} localize a single object per image just by exploiting a similarity graph at the level of an image; \cite{shin2022selfmask} proposes to combine different self-supervised representations, in an ensemble-fashion, and trains a model to learn the concept of object, the same as what~\cite{wang2022freesolo} does. However, most of these methods make assumptions about what an object is. For example, \cite{simeoni2021lost,wang2022tokencut} 
assume that an image contains more background pixels than object pixels, while
\cite{shin2022selfmask} discards masks that fill in the width of an image.
Such hypotheses restrict objects one can find.

In this work, we propose to tackle the problem the other way around: 
we make no assumptions about objects but focus instead on the concept of \emph{background}. Then, we use the idea that a pixel not belonging to the background is likely to belong to an object. 
Doing so, we do not need to make hypotheses about the number or the size of objects in order to find them. 
Our method, named \ours, is \emph{cheap} both at training and inference time. 

We start by computing a rough estimate of the background mask; this step works by mining a first patch that likely belongs to the background. 
To do this, we leverage attention maps in a self-supervised transformer and select one of the patches that received the least attention. Then the background mask incorporates patches similar to this mined one. 
One of our contributions is a reweighting scheme to reduce the effect of noisy attention maps based on the sparsity concept.
In the second step, we use the fact that the complement of this background mask provides an approximate estimation of the localization of the objects. 
This estimate is refined by training a single \conv layer on top of the frozen self-supervised transformer, using only the masks computed in the first step, an edge-preserving filter, and a self-labelling procedure. We show that this cheap method allows us to reach state-of-the-art results in the tasks of \emph{saliency detection}, \emph{unsupervised object discovery} and \emph{semantic segmentation retrieval}.

Our main contributions are as follows:
\begin{itemize}[leftmargin=15pt,topsep=0pt,itemsep=0pt,parsep=0pt,partopsep=0pt]
    \item We propose to think about the object discovery problem upside-down, and to look for what is \emph{not background} instead of directly looking for objects.
    \item We propose a new way to exploit already self-trained features and show that they allow us to discover the concept of background.
    \item We show that the use of attention heads can be improved by integrating a weighting scheme based on attention sparsity. 
    \item We propose a lightweight model composed only of a single \conv layer and show that there is no need to train a large segmenter for the task. 
    \item We demonstrate that our model performs well on \emph{unsupervised saliency detection}, \emph{unsupervised object discovery} and \emph{unsupervised semantic segmentation retrieval} tasks. We reach state-of-the-art results in all tasks with a method much faster and lighter than competing ones. 
\end{itemize}

\section{Related work}
\label{sec:related}

\paragraph{Self-supervised learning.} In self-supervised learning, a model is trained to solve a pretext task (e.g., jigsaw solving, colorization, or rotation prediction) on unlabeled data \cite{noroozi2016jigsaw,larsson2016colorization,gidaris2018rotnet,caron2021dino,gidaris2021obow,chen2020simclr,tian2020infomin,he2020moco,mocov2}.
Recently, with the surge of Vision Transformers (ViT) \cite{dosovitskiy2021vit} that stand out compared to convolutional networks, one can obtain rich, and dense descriptors of image patches with models trained in a self-supervised fashion on massive amounts of data \cite{caron2021dino,he2022mae,zhou2022ibot}.
For example, DINO~\cite{caron2021dino} employs a teacher-student framework where the two networks see different and randomly transformed input parts and the student network learns to predict the mean-centered output of the teacher network.
In MAE~\cite{he2022mae}, patches of the input image are randomly masked and the pretext task aims at learning to reconstruct the missing pixels by auto-encoding.
In these works, it has been shown that the representations of the self-attention maps of the ViTs contain interesting localization information
\cite{caron2021dino,he2022mae,zhou2022ibot,amir2021deep}, which have led recent methods to exploit these properties in several downstream tasks as unsupervised object discovery \cite{simeoni2021lost,wang2022tokencut,melas2022deepsectralmethod} or semantic segmentation \cite{hamilton2022stego,zadaianchuk2022comus,gansbeke2022maskdistill,vangansbeke2020unsupervised}.
In this paper, we build upon such self-supervised features to partition background and foreground patches. 
Arguably, learning self-supervised representation on unlabaled Imagenet~\cite{jia2009imagenet} --- a curated dataset --- induces a certain supervision. 
We leave for future work using models trained on less curated and more heterogenous datasets.

\paragraph{Unsupervised object localization.}
Localizing objects within images without any supervision is in the literature traditionally addressed by two distinct branches: 1)~unsupervised saliency detection methods find \emph{binary masks} of objects \cite{yan2013hierarchical_saliency,zhu2014saliency_optimization,li2015weighted_sparse_coding} while 2)~unsupervised object detection seeks for \emph{bounding boxes} around objects \cite{kim2009unsup_detection,jiang2013ufo,borji2014survey_salient_object_detection,zhu2014robust_background_detection}.
Unsupervised saliency detection has been approached with hand-crafted methods \cite{yan2013hs, zhu2014wctr}, generative adversarial models \cite{melas_kyriazi2021_imageseg}, or, closer to us, by refining noisy labels \cite{nguyen2019deepusps}. 
The first attempts in an unsupervised object discovery have often used region proposals \cite{zitnick2014edgebox, uijlings2013selectivesearch} as input. These works explored a collection of images and inter-image information using methods such as principal component analysis \cite{Wei2019ddtplus}, optimization \cite{vo2019unsup_image_matching,vo2020unsup_multi_object_discovery} or ranking \cite{vo2021largescale}.

Recently, these historically distinct tasks have been tackled jointly in unified frameworks \cite{wang2022tokencut, melas2022deepsectralmethod,ponimatkin2023videoobjectsegmentation} building on the advent of aforementioned self-trained dense visual features \cite{mocov2,caron2020swav,caron2021dino,he2022mae}. 
Given an image, these methods create a weighted graph where each node is a patch, and edges represent the similarity between the patches. Foreground objects are segmented by leveraging this similarity.
In particular, LOST \cite{simeoni2021lost} uses this graph to mine an object \textit{seed} as the patch with the least connection to other patches and expands the zone of interest to all connected similar patches afterwards.
Building on LOST, TokenCut~\cite{wang2022tokencut} and Deep Spectral Methods~\cite{melas2022deepsectralmethod} refine this result by using a normalized graph-cut to
separate an object from the highly connected patches, which most likely depict the background.

Another line of methods proposes to compute mask proposals that are later refined. SelfMask~\cite{shin2022selfmask} explores the use of multiple self-supervised features~\cite{mocov2,caron2020swav,caron2021dino} as the input of a spectral clustering algorithm.
FreeSOLO~\cite{wang2022freesolo} proposes FreeMask that generates correlation maps which are then ranked and filtered by a \emph{maskness} score.
DINOSAUR~\cite{seitzer2022dinosaur} performs representation learning by separating the features of an image and reconstructing them into individual objects or parts.

It should be noted that these prior works make strong underlying assumptions about what an object is.
This includes priors about the contrast~\cite{itti1998rapid_scene_analysis}, the size~\cite{simeoni2021lost,wang2022tokencut}, the centerness~\cite{judd2009humans_look}, the shape~\cite{shin2022selfmask} or boundary~\cite{wei2012geodesic} of the sought object. Instead, in our work, by looking for the background, we do not need to make any assumptions about the presence or number of objects. 

\paragraph{Learning to generalize through training.}
While we build our seed masks from single-image information, we refine these masks in a self-training step that leverages information shared across the whole image collection.
This self-training step aims at improving the quality of predictions by propagating and refining the initial seed of pseudo-annotations to a large set of unlabeled instances.
Early works in unsupervised saliency detection learn a deep unsupervised saliency network from noisy predictions obtained from handcrafted methods~\cite{zhang2017sup_by_fusion,zhang2018deep_usd,nguyen2019deepusps}. 
After clustering self-supervised features, \cite{simeoni2021lost,wang2022tokencut}~train a Class-Agnostic Detection (CAD) network over predicted pseudo-boxes and show that this trained detector can smooth out poor discoveries, therefore boosting results.
Similarly, in semantic segmentation, FreeSOLO \cite{wang2022freesolo} and COMUS \cite{zadaianchuk2022comus} feed coarse masks to train a segmentation model on these pseudo masks \cite{wang2020solo}. 

When propagating and refining pseudo-labels through the dataset with training, previous methods generally employ heavy training procedures involving learning several millions of parameters. Instead, our self-training step is \emph{extremely lightweight and fast} as it is only composed of one layer of 1$\times$1 convolutions and a two-epoch training scheme.


\section{Our method \ours}
\label{sec:method}

\begin{figure}
    \centering
    \includegraphics[width=\columnwidth,trim={0 0, 0.3cm 0},clip]{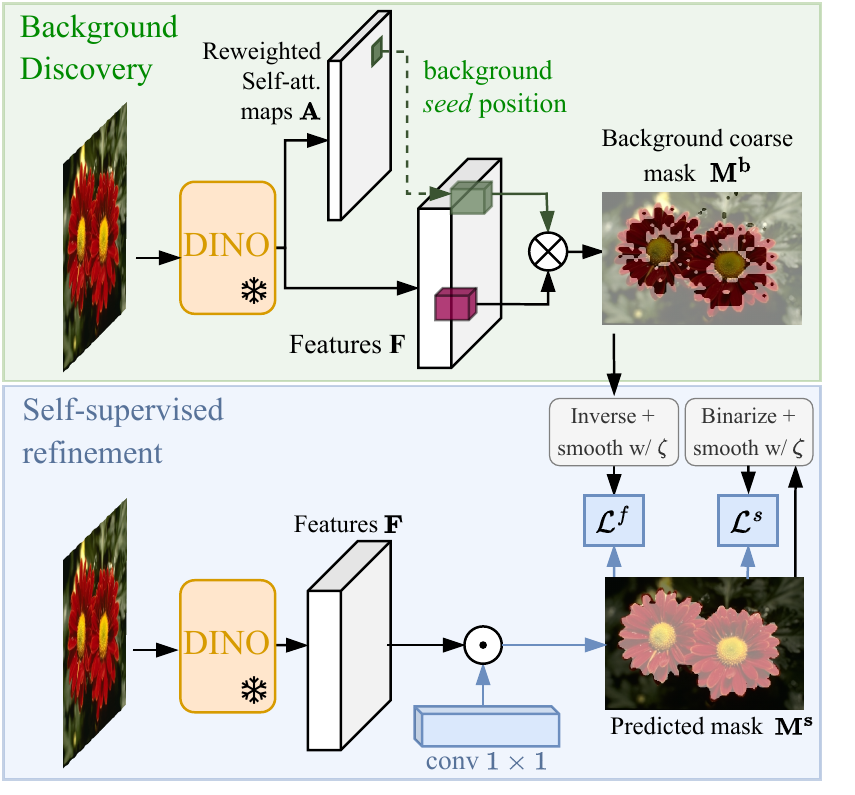}
    \caption{\textbf{Overview of \ours{}}.
    In the first stage (green upper part), a background mask $\mathbf{M^b}$ is discovered by mining a \emph{seed} patch through a reweighting of the self-attention maps of a frozen DINO \cite{caron2021dino}. This seed is then used to find similar patches likely belonging to the background.
    In the second stage (blue lower part), we train a lightweight $1\times1$ convolutional layer that produces refined masks from DINO features.
    It is trained in a self-supervised fashion to predict both smoothed inverse coarse masks $\mathbf{M^b}$ of the first step, and smoothed binarized version of its own output.
    Blue arrows denote where the gradients flow (in the reverse direction).
    }
    \vspace{-5pt}
    \label{fig:archi_found}
\end{figure}

In this work, we tackle the unsupervised object localization task by considering the problem upside-down.
Our approach consists of two stages.
First, we propose to look for patches corresponding to the \textit{background} in order to highlight patches that are likely objects (\autoref{sec:bkg-seg}).
Then, starting from these coarse masks, we design a fast and lightweight self-supervised learning scheme to refine them (\autoref{sec:bkg-learn}). 
An overview of \ours is shown in \autoref{fig:archi_found}.

\subsection{Background discovery}
\label{sec:bkg-seg}

Here, we look for the background pixels of an image $\mathbf{I} \in \mathbb{R}^{W \times H \times3}$.
To do so, we start by extracting deep features from this image using a self-supervised pre-trained ViT.
First, the image is divided into $N$ square patches of $P$ pixels each.
These patch tokens, along with an additional learned token, called class token (\cls), are processed by the ViT.
At the last self-attention layer, composed of $h$ different heads, we extract $h$ matrices $\{\mathbf{F}_i \in \mathbb{R}^{N \times d}\}_{i=1..h}$, that each contains $d$-dimensional features for each of the $N$ patches.
We also store in $\mathbf{A} \in \mathbb{R}^{N \times h}$ the $h$ self-attention maps between the \cls token and all patch tokens.

\paragraph{Background seed.} To identify the background, we start by identifying one patch which likely belongs to the background. This patch, called the background \textit{seed}, is defined as the patch with the least attention in $\mathbf{A}$ --- a patch which the model has learned to not give too much attention to. This seed is the $s^{\rm th}$ patch, where 
\begin{equation}
    s = \argmin_{p \ \in \ \{1, \ldots, N\}} \sum_{i=1}^{h}\mathbf{A}_{pi}.
    \label{eq:bkg-seed}
\end{equation}
In the equation above, $\mathbf{A}_{pi}$ is the attention score between the \cls token and the $p^{\rm th}$ patch in the $i^{\rm th}$ attention head.

\begin{figure}
    \centering
    \resizebox{\columnwidth}{!}{
     \renewcommand{\arraystretch}{.5}
     \begin{tabular}{cc}
        \centering
         \includegraphics[width=2.5cm]{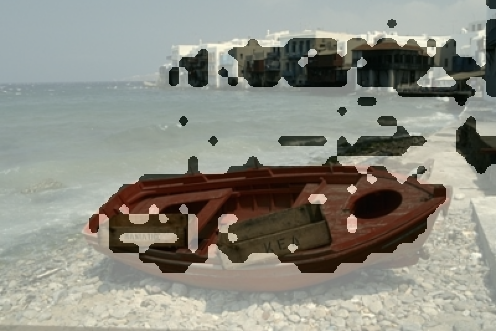} 
         &  
         \includegraphics[width=2.5cm]{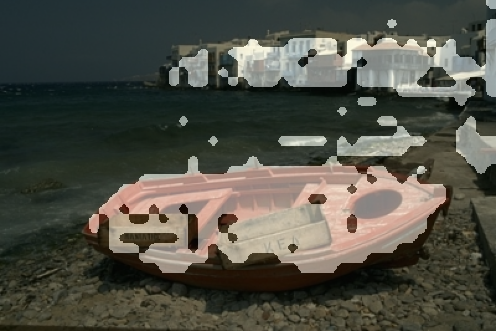} \\
         \tiny (a) Coarse background mask $\mathbf{M}^b$ & \tiny (b) Coarse foreground mask $\mathbf{M}^f$\\
         \includegraphics[width=2.5cm]{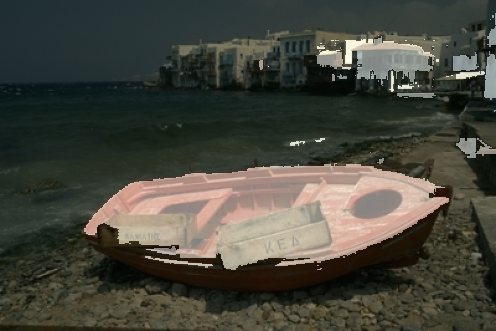} &
         \includegraphics[width=2.5cm]{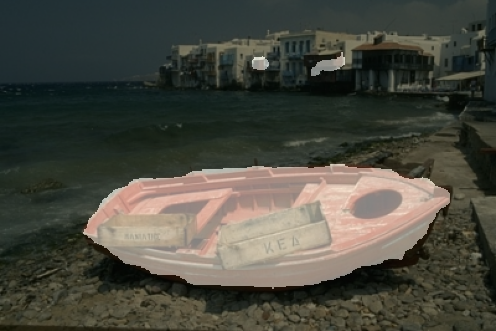} \\
         \tiny (c) Refined $\bs(\mathbf{M}^f)$ & \tiny (d) Predicted $\mathbf{M}^s$
    \end{tabular}}
    \caption{\textbf{Visualizations of masks generated for one image from ECSSD \cite{shi2016ecssd} at different stages of our method}. We show (a) the mask $\mathbf{M}^b$ extracted using our background discovery step, (b) its inverse $\mathbf{M}^f$ used as foreground mask to train our segmenter head, (c) the version refined using a bilateral solver $\zeta(\mathbf{M}^f)$ and (d) the output of our segmentation head $\mathbf{M}^s$ at the end of the training.
    }
    \label{fig:masks}
    \vspace{-5pt}
\end{figure}

\paragraph{Reweighting the attention heads.} When observing the $h$ different attention maps in $\mathbf{A}$, we notice that the background appears more or less clearly in the different heads. Therefore, we propose to weight each head differently in \autoref{eq:bkg-seed}. We exploit the sparsity of the attention map to compute these weights since the background appears better in a sparse attention map (as illustrated in the supplementary materials). Inspired by \cite{kalantidis2016crow}, we compute the 
 sparsity $S_i$ of each map by counting the number of attention values above a certain threshold $\mu > 0$: 
\begin{equation}
S_i = \vert \{p \ \vert \ A_{pi} \geq \mu, \; p = 1, \ldots, N \} \vert,
    \label{eq:mu}
\end{equation}
and reweight each attention map in \eqref{eq:bkg-seed} by
\begin{equation}
    w_i = \log \left(\frac{\sum_{j=1}^{h} S_j}{S_i} \right).
    \label{eq:w}
\end{equation}
Notice that $w_i$ increases when the sparsity $S_i$ decreases, i.e., when we visually observe a clearer separation of the background from the foreground thanks to sparser attention maps.
Finally, \autoref{eq:bkg-seed} becomes
\begin{equation}
    s = \argmin_{p \ \in \ \{1, \ldots, N\}} \sum_{i=1}^{h} w_i \, \mathbf{A}_{pi}.
    \label{eq:bkg-seed-w}
\end{equation}

\paragraph{Discovery of the background.} We identify the background by finding patches similar to the background seed. For each patch, we start by computing a single feature by concatenating the corresponding $d$-dimensional features in each head: ${\tilde{\mathbf{F}} = [w_1 \, \mathbf{F}_1, \ldots, w_h \, \mathbf{F}_h] \in \mathbb{R}^{N \times hd}}$. Then, the background mask $\mathbf{M}^b \in \{0, 1\}^N$ is defined as
\begin{equation}
    \mathbf{M}^b_p = 
    \begin{cases}
    1 & \text{if } {\rm sim}(\tilde{f}_p, \tilde{f}_s^{}) \geq \tau, \\
    0 & \text{otherwise},
    \end{cases}
    \; p=1, \ldots, N,
\end{equation}
for a threshold $\tau > 0$, where ${\rm sim}$ is the cosine similarity, and $\tilde{f}_p$, $\tilde{f}_s$ are the $hd$-dimensional features corresponding to patch $p$ and $s$ in $\tilde{\mathbf{F}}$. We show an example in \autoref{fig:masks}.

\subsection{Refining masks with self-training}
\label{sec:bkg-learn}

The proposed background discovery method described above is able to segment a good portion of the background but the corresponding masks are still far from perfect, as observed in \autoref{fig:masks}.
To improve them, and therefore to better segment the foreground objects, we propose a very \textit{simple} refinement step of learning a lightweight segmentation head in a self-supervised fashion.

The segmentation head consists of a single $1 \times 1$ convolution.
For each patch, it compresses DINO frozen patch features into a scalar, which is passed into a sigmoid function to encode the probability of the patch belonging to the foreground. 
We stress that, unlike most recent works, we do not train a heavy segmentation backbone~\cite{zadaianchuk2022comus,shin2022selfmask,wang2022freesolo} or a detection model~\cite{simeoni2021lost,wang2022tokencut}.
This aspect brings considerable training and inference efficiency both in terms of time and memory, as studied in \autoref{sec:costs}.

The segmentation head is trained in a self-supervised fashion. The general idea is that the model learns to predict a smoothed version of the complement of the coarse background masks and its own prediction, such that it quickly converges to refined masks.
We describe it formally below.

\paragraph{Segmentation head training.} 

Self-training is done thanks to two losses with distinct roles.
The first objective consists of initializing and guiding the predictions toward the coarse background masks.
The second objective aims at smoothing and refining predictions. 

Formally, let $\mathbf{M}^s \in \mathbb{R}^N$ be the soft output of the segmentation head for a given image.
The goal of the first objective is to predict the complement~$\mathbf{M}^f$ of the coarse background mask $\mathbf{M}^b$ (defined in \autoref{sec:bkg-seg}), refined by a bilateral solver $\bs(\cdot)$, which is an edge-aware smoothing technique improving the mask quality as proposed in~\cite{bilateralsolverbarron2016} and exploited in~\cite{wang2022tokencut,shin2022selfmask}.
Let $\hat{\mathbf{M}}^f \in  \{0,1\}^N$ be this refined version.
We compute the binary cross entropy 
\begin{equation}
   \mathcal{L}^f = \sum_{p=1}^N \left[ \hat{\mathbf{M}}^f_{p} \log \mathbf{M}^s_{p} + (1 - \hat{\mathbf{M}}^f_{p}) \log ( 1 - \mathbf{M}^s_{p}) \right]
   \label{eq:loss_bs_1}
\end{equation}
over a batch of images, where $\mathbf{M}^s_{p}$ and $\hat{\mathbf{M}}^f_{p}$ are the output of the segmentation head and refined coarse mask at patch $p$ respectively.
We additionally train the segmentation head by minimizing the binary cross entropy between the output and its refined version after binarization $\hat{\mathbf{M}}^s \in \mathbb{R}^N$, using again the bilateral solver, in order to force the quality of the mask edges, using
\begin{equation}
   \mathcal{L}^s = \sum_{p=1}^N \left[ \hat{\mathbf{M}}^s_{p} \log \mathbf{M}^s_{p} + (1 - \hat{\mathbf{M}}^s_{p}) \log ( 1 - \mathbf{M}^s_{p}) \right]
   \label{eq:loss_bs_2}
\end{equation}
with $\hat{\mathbf{M}}^s_{p}$ being the refined mask at patch $p$. Note that we compute this loss only for images for which $\mathbf{M}^s$ and $\hat{\mathbf{M}}^s$ do not differ too much, i.e., if ${\rm IoU}(\mathbf{M}^s, \hat{\mathbf{M}}^s) > 0.5$ following \cite{shin2022selfmask}. 
The two losses are linearly combined and balanced with a hyper-parameter $\lambda \in \mathbb{R}$: $\mathcal{L} = \mathcal{L}^f + \lambda \mathcal{L}^s$.
Also, after a few training steps, we observe that the model outputs become much better than the coarse masks.
Therefore, we stop using $\mathcal{L}^f$ after $m$ iterations. 
But, to avoid collapse we replace $\mathcal{L}^f$ with a cross-entropy loss that encourages predicted soft masks to be close to their binarized version.

\section{Experiments}
\label{sec:exp}
In this section, we make several experiments to assess the quality of \ours. We first evaluate it on the tasks of unsupervised object discovery (\autoref{sec:uod}), unsupervised saliency detection (\autoref{sec:saliency-detection}), and unsupervised semantic segmentation retrieval (\autoref{sec:retrieval}). Besides, we compare training/inference costs of the different methods in \autoref{sec:costs}, discuss qualitative results in \autoref{sec:qualitative}, and measure the impact of different components of our method in \autoref{sec:ablation}.

\paragraph{Technical details.}
In all experiments, we use a ViT-S/8 architecture \cite{dosovitskiy2021vit} pre-trained with \cite{caron2021dino}.
Following \cite{simeoni2021lost, wang2022tokencut}, we use the \texttt{key} features of the last attention layer as $\mathbf{F}$ and we use $\tau=0.3$ in the background discovery step.
The parameter $\mu$ in \autoref{eq:mu} is computed per image as the overall mean attention over all heads.
We use the coarse masks as pseudo ground-truth for $m=100$ iterations before refining the predictions directly. We balance the losses by setting $\lambda=1.5$.
Similar to \cite{shin2022selfmask}, we train \ours on DUTS-TR \cite{wang2017} (10,553 images) for $500$ iterations with a batch of $50$ images --- corresponding to a bit more than $2$ epochs.
We follow a similar training protocol as SelfMask \cite{shin2022selfmask}: we use random scaling with a range of $[0.1, 3.0]$ followed by image resizing to $(224,224)$ and Gaussian blurring applied with probability $0.5$. 
We use the parameters of the bilateral solver as provided by \cite{wang2022tokencut}.

In our evaluation, we consider two protocols: `\ours{} -- single' and `\ours{} -- multi'. In the `single' mode, we select the biggest connected component in $\mathbf{M}$. In the `multi' mode, we consider the mask as is --- with all detected objects. Additionally, when applying the bilateral solver $\bs()$, we extract, similarly, either the biggest connected component (single), or all connected components (multi). When not specified, we are using the `multi' setup.

\begin{table}
    \centering
    \resizebox{\linewidth}{!}{
    \begin{tabular}{lccc} \toprule
    Method & VOC07 & VOC12 & COCO20k \\
    \midrule
        \multicolumn{4}{c}{\bf--- \textit{No learning} ---} \\
        Selective Search \cite{uijlings2013selectivesearch} & 18.8 & 20.9 & 16.0 \\
        EdgeBoxes \cite{zitnick2014edgebox} & 31.1 & 31.6 &  28.8 \\
        Kim et al. \cite{kim2009unsup_detection} & 43.9 & 46.4 & 35.1 \\
        Zhang et al. \cite{zhang2020object} & 46.2 & 50.5 &  34.8 \\
        DDT+ \cite{Wei2019ddtplus} & 50.2 & 53.1 & 38.2 \\ 
        rOSD \cite{vo2020unsup_multi_object_discovery} & 54.5 & 55.3 & 48.5 \\
        LOD \cite{vo2021largescale} & 53.6 &  55.1 & 48.5 \\
        DINO-seg \cite{caron2021dino}\cite{simeoni2021lost} (ViT-S/16 \cite{caron2021dino}) & 45.8 & 46.2 & 42.0 \\
        LOST \cite{simeoni2021lost} (ViT-S/8 \cite{caron2021dino})  & 55.5 & 57.0 & 49.5 \\
        LOST \cite{simeoni2021lost} (ViT-S/16 \cite{caron2021dino}) & 61.9 & 64.0 & 50.7 \\
        DSS \cite{melas2022deepsectralmethod} (ViT-S/16 \cite{caron2021dino}) & 62.7 & 66.4 & 52.2 \\
        TokenCut \cite{wang2022tokencut} (ViT-S/8 \cite{caron2021dino}) \dag  & 67.3 & 71.6 &  60.7 \\
        TokenCut \cite{wang2022tokencut} (ViT-S/16 \cite{caron2021dino}) & 68.8 & 72.1 & 58.8 \\

        \multicolumn{4}{c}{\bf--- \textit{With learning} ---} \\
        FreeSolo \cite{wang2022freesolo} \dag  & 44.0 & 49.7 & 35.2 \\ 
        LOST + CAD \cite{simeoni2021lost} (ViT-S/16 \cite{caron2021dino}) & 65.7 & 70.4 & 57.5 \\
        TokenCut + CAD \cite{wang2022tokencut} (ViT-S/16 \cite{caron2021dino}) & 71.4 & \underline{75.3} & 62.6 \\
        SelfMask \cite{shin2022selfmask} \dag & \underline{72.3} & \underline{75.3} & 62.7 \\
        DINOSAUR \cite{seitzer2022dinosaur} & --- & 70.4 & \bf 67.2 \\
        \rowcolor{Apricot!40!white}
        \ours{} --- single (ViT-S/8 \cite{caron2021dino}) & \bf 72.5 & \bf 76.1 & \underline{62.9} \\ \bottomrule
        
    \end{tabular}
    }
    \caption{\textbf{Single object discovery results}. Comparative CorLoc performance on 3 datasets \cite{pascal-voc-2007, pascal-voc-2012, coco2014, vo2020unsup_multi_object_discovery}. `\dag': results from our own computation using TokenCut\cite{wang2022tokencut}, FreeSOLO \cite{wang2022freesolo} and SelfMask \cite{shin2022selfmask} available codes. `+CAD': a second-stage class-agnostic detector trained with unsupervised ``pseudo-boxes'' labels. All ViT backbones are trained following \cite{caron2021dino}. Best result is highlighted in \textbf{bold}, second best is \underline{underlined}.}
    \label{tab:obj_discovery}
\end{table}

\begin{table*}[ht]
    \centering
    \resizebox{0.90\textwidth}{!}{
    \begin{tabular}{lcccccccccc}
    \toprule
        & &  \multicolumn{3}{c}{DUT-OMRON\cite{yang2013saliency}} & \multicolumn{3}{c}{DUTS-TE \cite{wang2017}} & \multicolumn{3}{c}{ECSSD\cite{shi2016ecssd}} \\
        \cmidrule(rl){3-5}\cmidrule(rl){6-8}\cmidrule(rl){9-11}
         Method & Learning & Acc & IoU & max $F_\beta$ & Acc & IoU & max $F_\beta$ & Acc & IoU & max $F_\beta$ \\
        \midrule
        
        \multicolumn{11}{l}{\bf--- \textit{Without post-processing bilateral solver} ---} \\
         HS \cite{yan2013hs} &  & .843 & .433 & .561 & .826 & .369 & .504 & .847 & .508 & .673 \\
         wCtr \cite{zhu2014wctr} & & 838 & .416 & .541 & .835 & .392 & .522 & .862 & .517 & .684\\
         WSC \cite{li2015wsc}  & & .865 & .387 & .523 & .862 & .384 & .528 & .852 & .498 & .683\\
         DeepUSPS \cite{nguyen2019deepusps} & & .779 & .305 & .414 & .773 & .305 & .425 & .795 & .440 & .584 \\
         BigBiGAN \cite{voynov2021biggan}  & & .856 & .453 & .549 & .878 & .498 & .608 & .899 & .672 & .782 \\
         E-BigBiGAN \cite{voynov2021biggan}  & & .860 & .464 & .563 & .882 & .511 & .624 & .906 & .684 & .797 \\
         Melas-Kyriazi et al. \cite{melas2021}  & & .883 & .509 & --- & .893 & .528 & - & .915 & .713 & --- \\
         LOST\cite{simeoni2021lost} ViT-S/16 \cite{caron2021dino} & & .797 & .410 & .473 & .871 & .518 & .611 & .895 & .654 & .758 \\
         DSS \cite{melas2022deepsectralmethod}\cite{wang2022tokencut}& & --- & .567 & --- & --- & .514 & --- & --- & .733 & --- \\
         TokenCut\cite{wang2022tokencut} ViT-S/16 \cite{caron2021dino} & & .880 & .533 & .600 & .903 & .576 & .672 & .918 & .712 & .803 \\
         SelfMask\cite{shin2022selfmask} & \checkmark & .901 & \underline{.582} & --- & .923 & .626 & --- & \underline{.944} & .781 & --- \\
        \rowcolor{Apricot!40!white}
         \ours{} --- single ViT-S/8 \cite{caron2021dino} & \checkmark & \bf .920 & \bf .586 & \bf .683 & \bf .939 & \underline{.637} & \bf .733 & .912 & \underline{.793} & \underline{.946} \\ 
        \rowcolor{Apricot!40!white}
         \ours{} --- multi ViT-S/8 \cite{caron2021dino} &  \checkmark & \underline{.912} & .578 & \underline{.663} & \underline{.938} & \bf .645 & \underline{.715} & \bf .949  & \bf .807 & \bf .955 \\
         
        \multicolumn{11}{l}{\bf--- \textit{With post-processing bilateral solver} ---} \\
         LOST\cite{simeoni2021lost} ViT-S/16 \cite{caron2021dino} + $\zeta()$ & &.818 &.489 &.578 &.887 &.572 &.697 &.916 &.723 &.837 \\
         TokenCut\cite{wang2022tokencut} ViT-S/16 \cite{caron2021dino} + $\zeta()$  & & .897 & \underline{.618} & .697 &.914 &.624 & .755 &.934 &.772 &.874 \\
         SelfMask\cite{shin2022selfmask} + $\zeta()$ & \checkmark & .919 & \bf .655 & --- & .933 & \underline{.660} & --- & \bf.955 & \bf.818 & --- \\
        \rowcolor{Apricot!40!white}
         \ours{} --- single ViT-S/8 \cite{caron2021dino} + $\zeta()$ & \checkmark & \underline{.921} & .608 &  \underline{.706} & \underline{.941} & .654 & \underline{.760} & .949 & .805 & \underline{.934} \\ 
        \rowcolor{Apricot!40!white}
         \ours{} --- multi ViT-S/8 \cite{caron2021dino} + $\zeta()$ & \checkmark & \bf .922 & .613 & \bf .708 & \bf .942 & \bf .663 & \bf .763 & \underline{.951} & \underline{.813 }& \bf .935\\
         \bottomrule
        \end{tabular}}
    \caption{\textbf{Unsupervised saliency detection.} Performances of our method \ours \wrt state-of-the-art methods on the unsupervised saliency detection task. The symbol $\bs()$ denotes the application of the post-processing bilateral solver on the generated masks and the column `Learning' specifies which methods have a training step. We evaluate \ours in both the single and multi setup as described in main text. Best result per section is highlighted in \textbf{bold}, second best is \underline{underlined}.
    }
    \vspace{-5pt}
    \label{tab:saliency-detection}
\end{table*}

\subsection{Unsupervised object discovery}
\label{sec:uod}

We first evaluate our method on the task of unsupervised object discovery. We follow the common practice and use the \texttt{trainval} sets of PASCAL VOC07 \& VOC12 datasets \cite{pascal-voc-2007, pascal-voc-2012} and COCO20k (a subset of $19,817$ randomly chosen images from the COCO2014 trainval dataset \cite{coco2014} following \cite{vo2020unsup_multi_object_discovery, vo2021large_scale_unsup_object_discovery}). As in \cite{simeoni2021lost, wang2022tokencut, vo2021large_scale_unsup_object_discovery}, we report results with the Correct Localization (\textit{CorLoc}) metric. It measures the percentage of correct boxes, i.e., predicted boxes having an intersection-over-union greater than $0.5$ with one of the ground-truth boxes.

In \autoref{tab:obj_discovery}, we compare \ours{} -- single (no bilateral solver) to methods with no learning phase (LOST \cite{simeoni2021lost}, TokenCut \cite{wang2022tokencut}, DSS \cite{melas2022deepsectralmethod}), and to methods with a learning phase (SelfMask \cite{shin2022selfmask}, FreeSOLO \cite{wang2022freesolo}, and DINOSAUR \cite{seitzer2022dinosaur}). FreeSOLO \cite{wang2022freesolo} predicts multiple instance masks per image and, as such, we propose to merge all instances into a single mask, this gave us the best results. Other choices are discussed in the supplementary materials. For SelfMask \cite{shin2022selfmask}, if the mask contains multiple connected components, only the largest one is considered.

We show that \ours achieves state-of-the-art results on $2$ out of the $3$ datasets while being much cheaper to train. Indeed, the best method, DINOSAUR, achieves results significantly better than all others on COCO20k, but performs representation learning at a much higher training cost (as discussed in \autoref{sec:costs}). We note that it also achieves worse results than our method on VOC12 (-$5.7$pt). We discuss qualitative results in \autoref{sec:qualitative} and in the Supplemental.

\subsection{Unsupervised saliency detection}
\label{sec:saliency-detection}

We then consider the unsupervised saliency detection task, which is typically evaluated on a collection of datasets depicting a large variety of objects in different backgrounds. To compare to previous works, we evaluate on three popular saliency datasets: DUT-OMRON \cite{yang2013saliency} (5,168 images), DUTS-TE \cite{wang2017} (5,019 images), ECSSD \cite{shi2016ecssd} (1,000 images).
We report results in terms of intersection-over-union (\textit{IoU}), pixel accuracy (\textit{Acc}) and maximal $F_\beta$ score (max $F_\beta$) with $\beta^2=0.3$ following \cite{wang2022tokencut, shin2022selfmask} (additional details are given in the supplementary materials).
 
\autoref{tab:saliency-detection} presents our results compared to state-of-the-art methods, including LOST\cite{simeoni2021lost}, DeepSpectralMethods\cite{melas2022deepsectralmethod} (denoted DSS in the table), TokenCut \cite{wang2022tokencut} and the trained SelfMask \cite{shin2022selfmask}. 
When no bilateral solver is used, we observe that our method outperforms all methods, showing that we trained a good saliency estimator which produces high quality object masks.
With the application of the bilateral solver, we reach the same or better scores than the other methods, except for the IoU on DUT-OMRON. 
We observed that the bilateral solver sometimes amplifies the under-segmentation observed in the input mask (visual examples can be found in the supplementary materials). Correcting this behaviour is left for future work.

\begin{table}[ht!]
    \centering
    \resizebox{0.81\linewidth}{!}{
    \begin{tabular}{lcc} 
    \toprule
    \multicolumn{1}{c}{} & \multicolumn{2}{c}{mIoU} \\ \cmidrule{2-3}
    Method & 7cls & 21cls \\
    \midrule
    \multicolumn{3}{c}{\bf--- \textit{Representation learning methods} --- } \\
    MaskContrast\cite{vangansbeke2020unsupervised} (unsup. sal.) $\circ$ & 53.4 & 43.3 \\
    
    \multicolumn{3}{c}{\bf--- \textit{Single saliency mask} ---} \\
    FreeSOLO \cite{wang2022freesolo} & 19.7 & 17.0 \\
    FreeSOLO \cite{wang2022freesolo} (largest inst.) & 20.6 & 20.6 \\
    TokenCut \cite{wang2022tokencut} (ViT-S/8 \cite{caron2021dino}) & 46.7 & 37.6 \\
    TokenCut \cite{wang2022tokencut} (ViT-S/16 \cite{caron2021dino}) & 49.7 & 39.9 \\
    SelfMask \cite{shin2022selfmask} & 56.6 & 40.7 \\
    \rowcolor{Apricot!40!white}
    \ours (ViT-S/8 \cite{caron2021dino}) & 56.1 & \bf 42.9 \\
    
    \multicolumn{3}{c}{\bf--- \textit{Single saliency mask + bilateral solver} ---} \\
    FreeSOLO \cite{wang2022freesolo} & 20.2 & 17.3 \\
    TokenCut \cite{wang2022tokencut} (ViT-S/8 \cite{caron2021dino}) & 47.2 & 37.2 \\
    TokenCut \cite{wang2022tokencut} (ViT-S/16 \cite{caron2021dino}) & 50.2 & 39.8 \\
    SelfMask \cite{shin2022selfmask} & 55.4 & 40.9 \\
    \rowcolor{Apricot!40!white}
    \ours (ViT-S/8 \cite{caron2021dino}) & \underline{57.2} & 42.2 \\
    
    \multicolumn{3}{c}{\bf--- \textit{Multiple saliency masks} ---} \\
    FreeSOLO \cite{wang2022freesolo} & 23.9 & 25.7 \\
    SelfMask \cite{shin2022selfmask} & 56.2 & 40.8 \\
    \rowcolor{Apricot!40!white}
    \ours (ViT-S/8 \cite{caron2021dino}) & \textbf{58.0} & \underline{42.7} \\
    
    \bottomrule
    
    \end{tabular}
    }
    \caption{\textbf{Retrieval on PASCAL VOC12 \cite{pascal-voc-2012}.} 
    We consider either a single instance per image 
    (the second and the third blocks in the table)
    or multiple instances in each image 
    (last block). 
    Feature extractor used to get saliency prediction in \ours, TokenCut, and SelfMask is indicated between parentheses.
    All methods except MaskContrast use features from \texttt{ViT-S/8} during retrieval.
    Best result is highlighted in \textbf{bold}, second best is \underline{underlined}.
    $\circ$ denotes result reported from~\cite{vangansbeke2020unsupervised}. 
    }
    \label{tab:retrieval}
\end{table}

\subsection{Semantic Segmentation Retrieval}
\label{sec:retrieval}
In this section, we test our method on the task of unsupervised semantic segmentation retrieval on the PASCAL VOC12 \cite{pascal-voc-2012} dataset in order to evaluate the quality of the predicted saliency masks. We follow a protocol proposed by \cite{vangansbeke2020unsupervised} and compare to related methods whose code is available online, namely to TokenCut~\cite{wang2022tokencut}, SelfMask~\cite{shin2022selfmask} and FreeSOLO~\cite{wang2022freesolo}.
We also include a comparison to MaskContrast~\cite{vangansbeke2020unsupervised}, which takes the opposite approach to ours as it trains the feature representations while having a frozen pre-trained saliency predictor. 
We consider two different evaluation setups. First, \textbf{(a)} we assume that the predicted mask depicts a single object. 
For FreeSOLO~\cite{wang2022freesolo}, which generates several instances per image, we tried several combinations and merged all instances into a single one or consider only the largest instance (noted ``\emph{largest inst.}'').
\textbf{(b)}~We test the multiple-instances setting, which is more fair to FreeSOLO, and allows us to evaluate the ability of \ours to separate objects. In this setup, we consider each instance of FreeSOLO as an object. For all other methods, we compute the connected components in the mask outputs, and each component is then treated as an object (we discard those smaller than $1\%$ of an input image size).

Given an object mask, we compute a per-object feature vector averaged over the corresponding pixels.
We apply this procedure both in the train and val splits. 
We use a ViT-S/8 trained using DINO~\cite{caron2021dino} as a feature extractor for \ours, TokenCut, SelfMask, and FreeSOLO. MaskContrast uses its own \emph{optimized} feature extractor.
Finally, we find the nearest neighbors of each object of the val set to objects in the train set and assign them the corresponding ground-truth label. We measure the mean Intersection-over-Union (\textit{mIoU}) between the predictions and ground truths.

Results in \autoref{tab:retrieval} are given for both setups and are computed either over $7$ (bus, airplane, car, person, cat, cow and bottle) or all $21$ classes of the VOC dataset, following~\cite{vangansbeke2020unsupervised}.
We can observe that \ours outperforms all methods in both cases by a consistent margin. Results also confirm SelfMask as a strong competitor that is however outperformed by \ours across all considered setups with gaps between 1.3 and 2.2 mIoU points, excepting the single saliency with 7 classes evaluation where SelfMask surpasses \ours by 0.5 point.
Improvements of \ours over TokenCut and FreeSOLO can be explained because TokenCut localizes only a single object per image and FreeSOLO finds objects that are often not considered as so in the dataset.
We continue the discussion in \autoref{sec:qualitative}.

{\setlength{\tabcolsep}{0.5pt}
\begin{figure*}[t!]
    \small
    \centering
    \resizebox{\textwidth}{!}{
    \begin{tabular}{cccccc}
     (a) Input image & 
     (b) Ground truth & 
     (c) \ours (ours) & 
     (d) TokenCut~\cite{wang2022tokencut} & 
     (e) SelfMask~\cite{shin2022selfmask} & 
     (f) FreeSOLO~\cite{wang2022freesolo} \\
    \includegraphics[width=3.0cm,height=1.5cm]{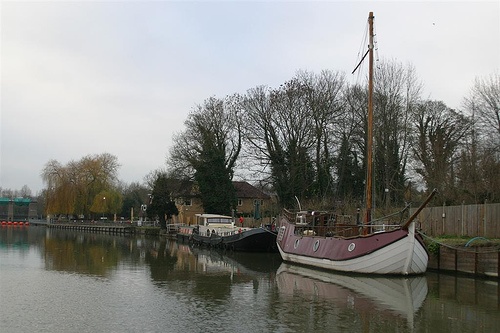} & 
    \includegraphics[width=3.0cm,height=1.5cm]{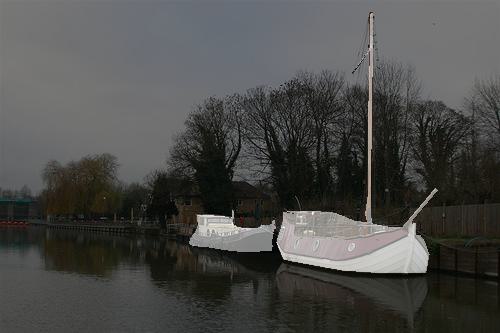} & 
    \includegraphics[width=3.0cm,height=1.5cm]{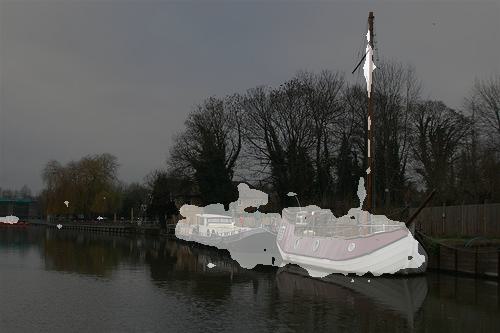} & 
    \includegraphics[width=3.0cm,height=1.5cm]{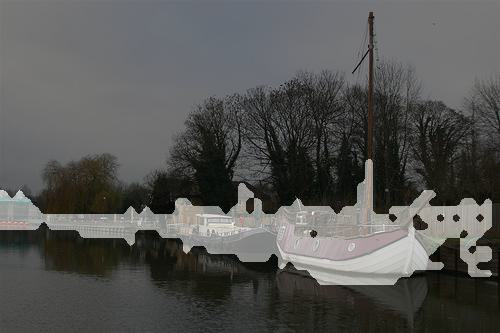} & 
    \includegraphics[width=3.0cm,height=1.5cm]{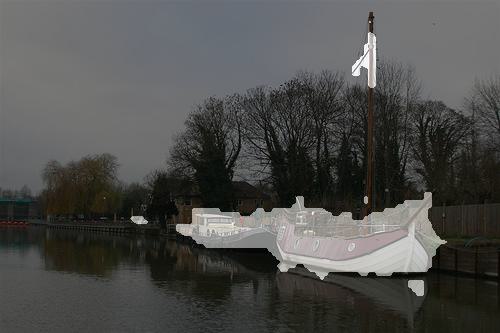} & 
    \includegraphics[width=3.0cm,height=1.5cm]{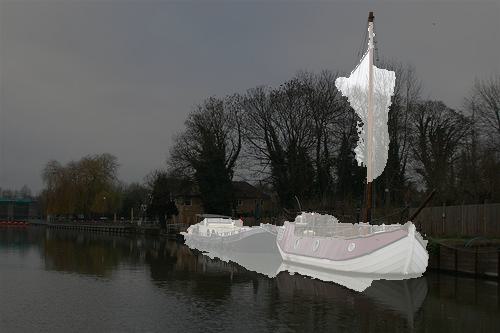} \vspace{-2pt} \\

    \includegraphics[width=3.0cm,height=1.5cm]{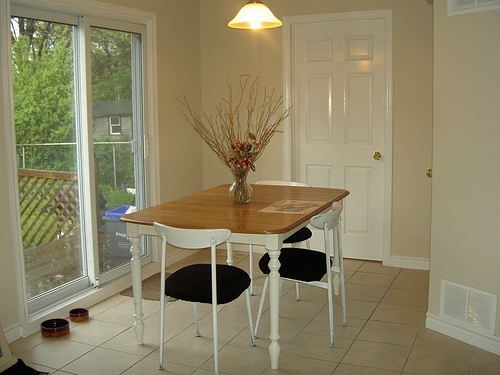} & 
    \includegraphics[width=3.0cm,height=1.5cm]{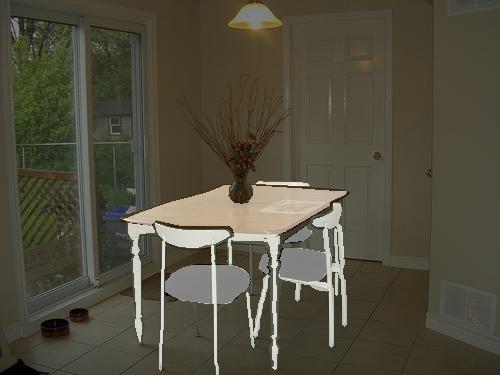} & 
    \includegraphics[width=3.0cm,height=1.5cm]{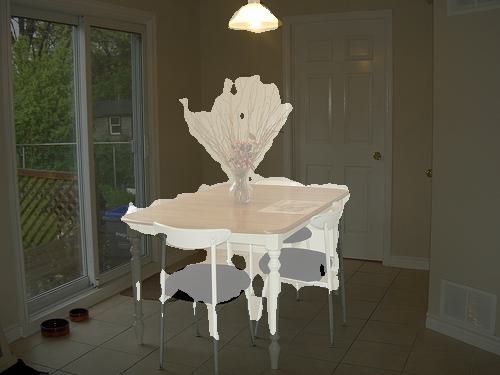} & 
    \includegraphics[width=3.0cm,height=1.5cm]{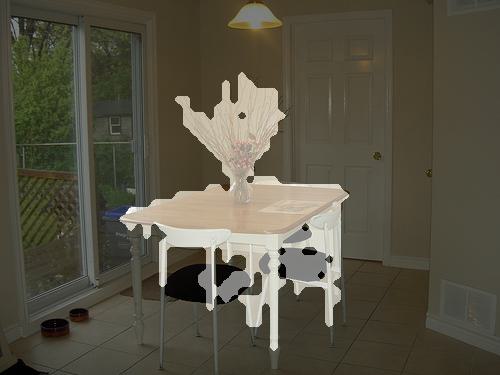} & 
    \includegraphics[width=3.0cm,height=1.5cm]{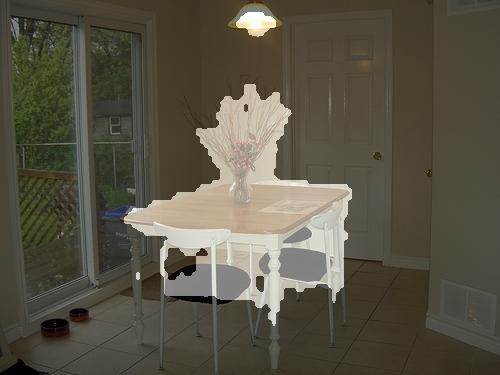} & 
    \includegraphics[width=3.0cm,height=1.5cm]{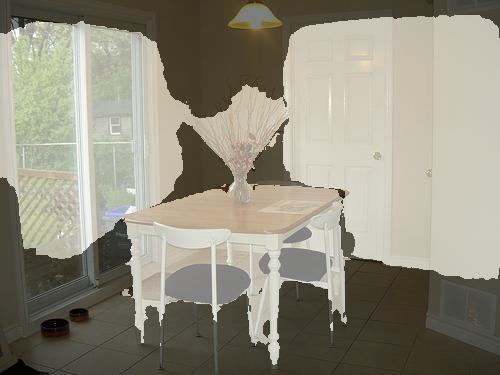}
    \vspace{-2pt} \\
    
    \includegraphics[width=3.0cm,height=1.5cm]{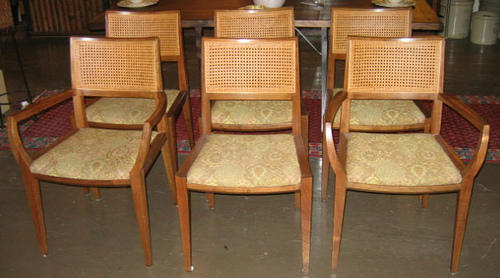} &
    \includegraphics[width=3.0cm,height=1.5cm]{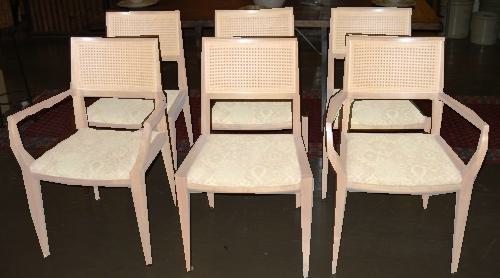} & 
    \includegraphics[width=3.0cm,height=1.5cm]{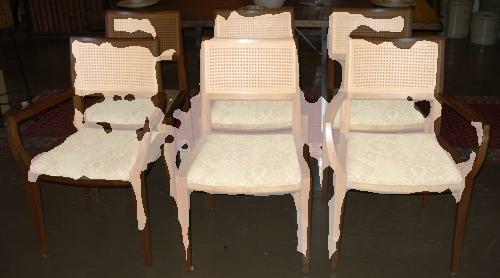} & 
    \includegraphics[width=3.0cm,height=1.5cm]{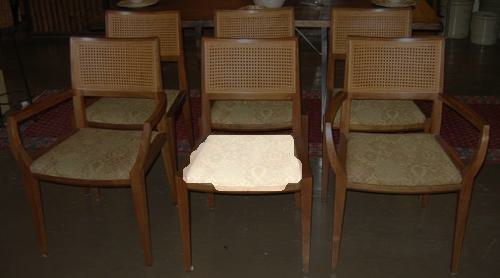} & \includegraphics[width=3.0cm,height=1.5cm]{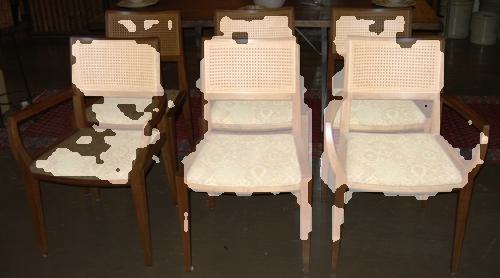} & \includegraphics[width=3.0cm,height=1.5cm]{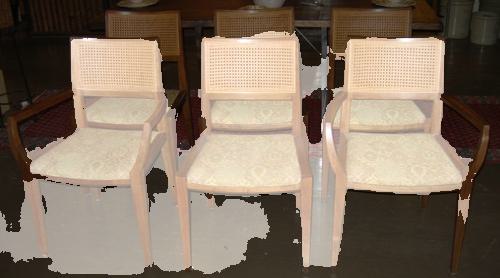} 
    \end{tabular}
    }
    \vspace{-5pt}
    \caption{\textbf{Qualitative results of object localization.}
    We overlay predicted masks generated with our method \ours, TokenCut~\cite{wang2022tokencut}, SelfMask~\cite{shin2022selfmask} and FreeSOLO~\cite{wang2022freesolo} on three images taken from VOC12 \cite{pascal-voc-2012}.}
    \label{fig:qual_v2}
\end{figure*}
}

\begin{table}[t]
\centering
\resizebox{.92\linewidth}{!}{
\begin{tabular}{lcc} \toprule
 & $\#$ learnable & inference \\
Method &  params. & FPS\\
\midrule
LOST \cite{simeoni2021lost} & --- & 64 \\
TokenCut \cite{wang2022tokencut} & --- & 0.4 \\
SelfMask \cite{shin2022selfmask} & $\approx 36$M & 13 \\
FreeSOLO \cite{wang2022freesolo} & $\approx 66$M & 13 \\
DINOSAUR \cite{seitzer2022dinosaur} -- MLP dec. $\ast$ & $\geq$ $5$M  & --- \\ 
DINOSAUR \cite{seitzer2022dinosaur} -- transf. dec.$\ast$  & $\geq$ $77$M  & --- \\
\rowcolor{Apricot!40!white}
\ours & 770 & 80 \\ \bottomrule                               
\end{tabular}
}
\caption{\textbf{Memory and inference costs.} Comparison of the cost of the different methods.
`$\#$ learnable params.' excludes weights of the frozen DINO backbone.
The FPS measure includes the forward pass through DINO
and is computed on a single V100 GPU with PyTorch 1.8.1.
`$\ast$':  denotes an estimation of the number of learnable parameters for methods without public code. 
} 
\label{tab:nb_params_comp}
\end{table}

\subsection{Comparison of method costs}
\label{sec:costs}
We compare \ours to methods that either do or do not include training, and that have very different costs at inference time. In this section, we highlight the advantage of our method in terms of complexity and speed. 
\ours is a segmenter head composed of just $770$ parameters, trained over $2$ epochs on DUTS-TR \cite{wang2017} on a single GPU, and which can infer at \emph{$80$ FPS}, including the forward pass through DINO, on a V100 GPU.
We summarize key numbers in \autoref{tab:nb_params_comp}. 

First, regarding methods with no training, \cite{wang2022tokencut} 
requires the costly computation of an eigenvector on the Laplacian matrix of the affinity graph, therefore making the method rather slow (0.4 FPS). 
For the same reasons, \cite{melas2022deepsectralmethod} runs at equivalent speed to \cite{wang2022tokencut}. 
LOST~\cite{simeoni2021lost} is almost as fast as us but achieves much lower performance, as seen before.

Second, regarding methods that include training, SelfMask~\cite{shin2022selfmask} trains a model of $\approx36$M parameters over $12$ epochs on DUTS-TR~\cite{wang2017}, by exploiting $27$ mask proposals generated using three different backbones, thus making the training considerably more expensive than ours.
FreeSOLO~\cite{wang2022freesolo} proposes a faster mask proposal extraction step using a DenseCL~\cite{wang2021dense} model based on a ResNet~\cite{he2016residual} backbone. It then trains a SOLO~\cite{wang2020solo} model ($\approx 65$M learnable parameters) for in total $60$k iterations on 8 GPUs, making it much more expensive to train compared to us.

\subsection{Qualitative results}
\label{sec:qualitative}

We show visualizations of saliency masks predicted by \ours and related methods in \autoref{fig:qual_v2}.
We notice that FreeSolo \cite{wang2022freesolo} and SelfMask \cite{shin2022selfmask} tend to oversegment the objects in all examples, while \ours yields masks much more accurate with respect to the ground truth. Regarding TokenCut \cite{wang2022tokencut}, we observe, in the last row of the figure, that it segments just a part of one chair, while \ours segments all the chair rather accurately. These examples illustrate the efficiency of our method in dealing with multiple objects. 

\subsection{Ablation study}
\label{sec:ablation}
We present in \autoref{tab:ablation} an ablation study of 
our method on the saliency dataset ECSSD~\cite{shi2016ecssd} --- more can be found in Appendix. We measure scores on the unsupervised saliency detection task following the protocol detailed in~\autoref{sec:saliency-detection}. 

\paragraph{Coarse masks.} 
We evaluate our background discovery method (\autoref{sec:bkg-seg}) with and without the attention head reweighting scheme (column R in~\autoref{tab:ablation}). We can observe that the reweighting boosts results up to $1$pt when evaluated in a multi-setup mode. We also compare results with and without the application of the post-processing bilateral solver, noted $\bs()_p$, and observe that the refined masks yield better results by $3$pts of IoU in the ``single'' setting. Such improvements (visualized in \autoref{fig:masks} and the supplementary materials) are significant.
Overall, our background discovery method (\autoref{sec:bkg-seg}) already achieves decent results, particularly when considering the \emph{single} setup. As discussed before and observed in~\autoref{fig:masks}, our coarse maps cover several objects and do not focus only on the most salient one.

\begin{table}[ht!]
    \centering
    \resizebox{0.95\linewidth}{!}{
    \begin{tabular}{lcccccc}
    \toprule
    Method & R & $\zeta()_t$ & $\zeta()_p$ & Acc & IoU & max $F_\beta$ \\
    \midrule
        \multicolumn{7}{c}{\bf --- \textit{Coarse masks, no training} ---} \\
        \autoref{sec:bkg-seg} -- multi &  & \cellcolor{gray!15} &  & .876 & .627 & .689 \\
        \autoref{sec:bkg-seg} -- multi & \checkmark  & \cellcolor{gray!15} & & .880 & .637 & .702 \\
        \autoref{sec:bkg-seg} -- single &  &  \cellcolor{gray!15} & & .898 & .671  & .746 \\
        \autoref{sec:bkg-seg} -- single & \checkmark & \cellcolor{gray!15} &  & .901 & .679 & .758 \\
        \autoref{sec:bkg-seg} -- single & & \cellcolor{gray!15} & \checkmark & .906 & .709 & .780 \\
        \autoref{sec:bkg-seg} -- single & \checkmark & \cellcolor{gray!15} & \checkmark & \bf .909 & \bf .717 & \bf .792\\

        \multicolumn{7}{c}{\bf--- \textit{With training} ---} \\
        \ours{} -- multi & \checkmark &  &  & .944 & .790 & .886\\
        \rowcolor{Apricot!40!white}
        \ours{}  -- multi & \checkmark & \checkmark &  &.949 &.807 & \bf .955 \\
        \rowcolor{Apricot!40!white}
        \ours{}  -- multi & \checkmark & \checkmark & \checkmark & \bf .951 & \bf .813 &.935 \\
    \bottomrule
    \end{tabular}}
    \caption{\textbf{Ablation study.} Study of the impact of the different elements in the background discovery step (\autoref{sec:bkg-seg}). Results are provided following the unsupervised saliency detection protocol on the ECSSD \cite{shi2016ecssd} dataset. $R$ stands for the reweighting of the attention heads. 
    We note $\zeta()_t$ and $\zeta()_p$ the application of the bilateral solver during training (\autoref{eq:loss_bs_1}-\ref{eq:loss_bs_2}) and as post-processing.
    }
    \label{tab:ablation}
\end{table}

\paragraph{The impact of learning} In the same table, we present results obtained after the training of the single \conv layer. Training over coarse masks provides a significant boost of more than $15$ IoU pts in the multi setup.
This shows that the model learns the concept of foreground objects and smooth results over the dataset.
Using the bilateral solver in \autoref{eq:loss_bs_1}-\ref{eq:loss_bs_2}, noted $\bs()_t$, further improves results by 1.7 IoU pts and by an additional .6 pts when also applied as post-processing.

\section{Discussion}
In this work, we address the problem of unsupervised object localization, that we propose to attack sideways: we look first for the scene background --- using self-supervised features --- instead of looking for the objects directly.
Putting this simple idea at work, we extract coarse masks that encompass most of the background, their complements thus highlighting objects.
Using the inverse of the background masks,
we train a lightweight segmenter head made of only $770$ learned parameters, which runs at $80$ FPS at inference time --- including the forward pass through the backbone --- and reaches state-of-the-art results in unsupervised object discovery, unsupervised saliency detection, and unsupervised instance segmentation retrieval. 

\vspace{-3pt}
\paragraph{Acknowledgments}
This work was supported by the HPC resources of GENCI-IDRIS in France under the 2021
grant AD011013413, 
and 
by the ANR grant MultiTrans (ANR-21-CE23-0032),
It was also supported 
by the Ministry of Education, Youth and Sports of the Czech Republic through the e-INFRA CZ (ID:90140) and by CTU Student Grant SGS21\/184\/OHK3\/3T\/37. 

{\small
\bibliographystyle{ieee_fullname}
\bibliography{egbib}
}

\appendix
\newpage

\setcounter{figure}{4}
\setcounter{table}{5}
\setlength{\tabcolsep}{1pt}

\section{Extra details}
\subsection{During learning}

During training, $\bs()$ is applied at the image resolution.
To do so, masks are upsampled to the original image size and the output refined masks are downsampled to the feature map size. 
The model is trained with the AdamW optimizer provided by PyTorch, with an initial learning rate of $5\text{e}\!-\!2$. We use a simple step scheduler which applies a decay of $0.95$ every $50$ iterations.

\subsection{Unsupervised saliency detection}
We detail here the different metrics used in the task of unsupervised saliency detection.

\paragraph{The maximal $F_\beta$ metric\!\!\!} is the maximum $F_\beta$ over various masks which have been binarized using different thresholds. Formally, $F_\beta$ is the harmonic mean of precision (P) and recall (R) between a binary mask $M$ and the ground-truth mask $G$, i.e.,
\begin{equation}
    F_\beta = \frac{(1 + \beta^2) \; \text{P} \times \text{R}}{ \beta^2\;\text{P} + \text{R}},
\end{equation}
where $\beta^2$ is the precision weight, set at $0.3$ following \cite{shin2022selfmask, liu2019poolingbased, melas_kyriazi2021_imageseg, zhao2019_edgnet}. The max $F_\beta$ is computed by taking a soft predicted mask $M_p \in [0, 255]$ and binarizing it using $255$ different thresholds between $0$ and $254$; max $F_\beta$ is then the maximum value of $F_\beta$ among all the generated binary masks, taken over the whole dataset (single optimal threshold). 
We noticed in SelfMask's code that the maximal $F_\beta$ is computed with an optimal threshold found for each image rather than over the whole dataset. For this reason, and for a fair comparison, we do not report this original max $F_\beta$ in our unsupervised saliency detection table.

\paragraph{The Intersection-over-Union\!\!\!} measures the overlap between foreground regions of a predicted binary mask and the ground-truth mask, averaged over the entire dataset.

\paragraph{The pixel accuracy metric\!\!\!} measures the pixel-wise accuracy between a predicted binary mask $M \in \{0,1\} ^ {H\times W}$ and the corresponding ground-truth mask $G \in \{0,1\} ^ {H\times W}$. Formally, it can be defined as:
\begin{equation}
    \text { Acc }=\frac{1}{H \times W} \sum_{i=1}^H \sum_{j=1}^W \delta_{G_{i j}, M_{i j}},
\end{equation}
with $\delta$ being the Kroneker-delta function and $G_{i j}$, $M_{i j}$ being the value of the ground-truth and predicted masks at position $(i,j) \in \{1\cdots H\}\times\{1\ldots W\}$.

\subsection{Different setups for FreeSOLO}
\label{sec:sup-freesolo}

FreeSOLO~\cite{wang2022freesolo} is a class-agnostic instance segmentation method and outputs several instance masks per image, making
it different to other baselines. 
In order to compare it to our method, we use the code provided online.
We follow the original paper to get the prediction masks,
i.e., we apply matrix non-maximum suppresion (NMS)~\cite{wang2020solo}
and keep masks with a maskness score above $0.7$.

\paragraph{Unsupervised object discovery} We present in \textcolor{red}{Sec. 4.1} of the main paper our unsupervised object discovery protocol.
The extraction of the single object box is straightforward for all methods but FreeSOLO \cite{wang2022freesolo}. 
For this method we have considered three setups:
(a) merging all instance masks into a single one;
(b) keeping only the mask with the highest maskness score; 
(c) keeping only the mask containing the largest connected component. Best results were achieved with (a) and are reported in the main paper.

\paragraph{Semantic segmentation retrieval}
We have performed similar tests with FreeSOLO \cite{wang2022freesolo} in the semantic segmentation retrieval task. 
Additionally to the evaluation setups described in the main paper, we have experimented using two or more instances but without improvements of the results. 

\subsection{Semantic segmentation retrieval}
In the task of unsupervised semantic segmentation retrieval, we consider two setups.
One considers that the predicted mask highlights a single object, while the other splits the mask into connected components and treats each component
as individual object.
In both cases, we compute a per-object feature vector averaged over the pixels of the considered mask. Given a (flattened) binary mask $M \in \{0,1\}^{H W\times 1}$ and corresponding feature tensor $F \in \mathbb{R}^{C\times HW}$ with $C$ the number of channels, we obtain a prototype $P\in\mathbb{R}^{C}$ as
\begin{equation}
    P = F M.
\end{equation}
These prototypes are first extracted for all train samples and serve as an index for retrieval. Then, to get a label for each val sample, we compute the sample
prototype, find nearest neighbors in the train prototypes, and assign it the corresponding label.

\begin{figure}[ht!]
\resizebox{\columnwidth}{!}{
\begin{tabular}{cc}
    \centering
    \begin{tikzpicture}
    \tikzstyle{every node}=[font=\small]
      \begin{axis}[ 
          width=4cm,
          height=3.5cm,
          line width=0.5,
          grid=major, 
          grid style={white},
          xlabel={values of $\tau$},
          ylabel={IoU},
          ytick pos=left,
          title={\small DUT-OMRON},
          xtick={0.1,0.3,0.5},
          xticklabels={0.1,0.3,0.5},
          xtick pos=bottom,
        legend style={anchor=south, draw=none, fill=none},
        ymin = 58.1,
        ymax = 58.8,
        legend pos=south west,
        grid=major,
        legend cell align={left},
          ]
        \addplot[orange,mark=+] coordinates {(0.1,58.6) (0.2,58.6) (0.3,58.6) (0.4,58.5) (0.5,58.4)}; 
      \end{axis}
    \end{tikzpicture} 
    & 
    
    \begin{tikzpicture}
    \tikzstyle{every node}=[font=\small]
      \begin{axis}[ 
          width=4cm,
          height=3.5cm,
          line width=0.5,
          grid=major, 
          grid style={white},
          xlabel={values of $\tau$},
          title={\small ECSSD},
          ytick pos=left,   
          xtick={0.1,0.3,0.5},
          xticklabels={0.1,0.3,0.5},
          xtick pos=bottom,
          legend style={anchor=south, draw=none, fill=none},
          ymin = 79.,
          ymax = 79.6,
          legend pos=south west,
          legend cell align={left},
          ]
        \addplot[color=purple,mark=+] coordinates {(0.1,79.3) (0.2,79.3) (0.3,79.3) (0.4,79.3) (0.5,79.3)};
      \end{axis}
    \end{tikzpicture}
\end{tabular}
}
\caption{\textbf{Sensitivity to background threshold $\tau$}.
We report saliency detection results measured on the datasets DUT-OMRON (\textit{left}) and ECSSD (\textit{right}) with the IoU metric.}
\label{fig:tau}
\end{figure}
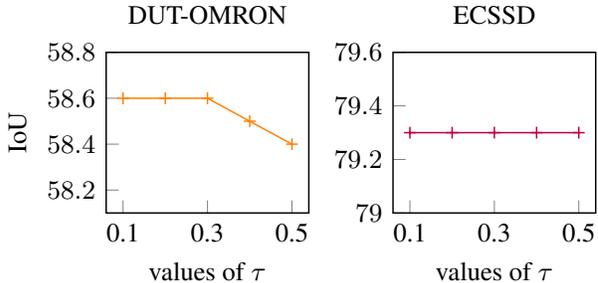

\section{Sensitivity to masking method}
\subsection{Sensitivity to background threshold $\tau$}
We investigate here the impact of the background parameter $\tau$ on final results. We report in \autoref{fig:tau} saliency detection results. We observe that \ours is stable to changes of 
$\tau \!\in\! [0.1,0.5]$, with saliency scores varying by at most 0.2 percentage pts on DUT-OMRON and not at all on ECSSD.

\setlength{\tabcolsep}{5pt}
\begin{table}[]
    \centering
    
    \begin{tabular}{l|ccc}
         method & VOC07 & VOC12 & COCO20k \\
         \toprule
         TokenCut \cite{wang2022tokencut} + T & 72.3 & 75.9 & 62.7 \\
         LOST \cite{simeoni2021lost} + T & 72.3 & \bf 76.1 & 62.8 \\
         \ours (ours) &\bf 72.5 & \bf 76.1 & \bf 62.9 \\
         
    \end{tabular}
    \caption{\textbf{Sensitivity to mask generation method.}
    Unsupervised object discovery results (measured using the CorLoc metric) when using different mask generation strategy to generate the masks $\mathbf{M}^f$ refined in our training process. T denotes the training of our segmentation head with the masks $\mathbf{M}^f$.}
    \label{tab:coarse-masks}
\end{table}
\setlength{\tabcolsep}{1pt}

\begin{figure*}[ht!]
\centering
    \includegraphics[height=2.6cm]{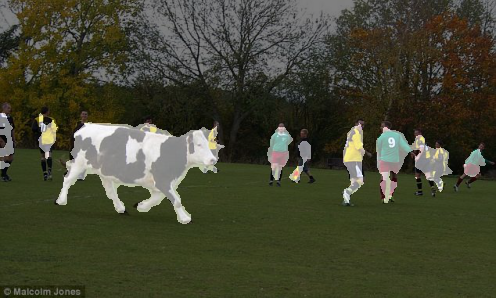}
    \includegraphics[height=2.6cm, trim={1cm 0 1cm 0},clip]{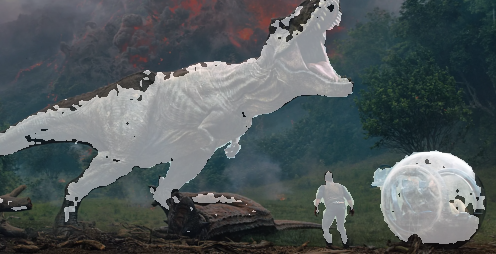}
    \includegraphics[height=2.6cm, trim={4cm 0 0cm 0},clip]{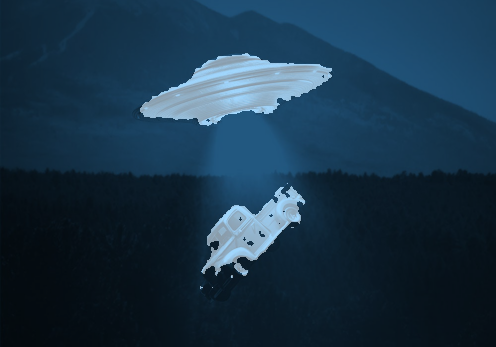}
    \includegraphics[height=2.6cm]{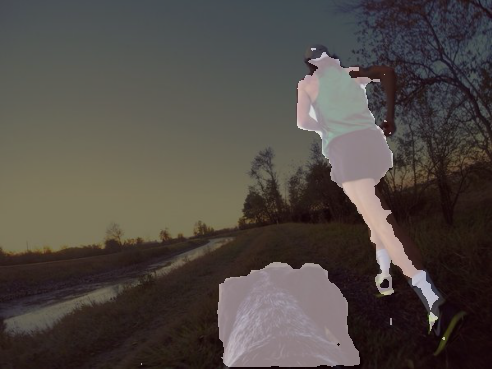}
    \includegraphics[height=2.6cm]{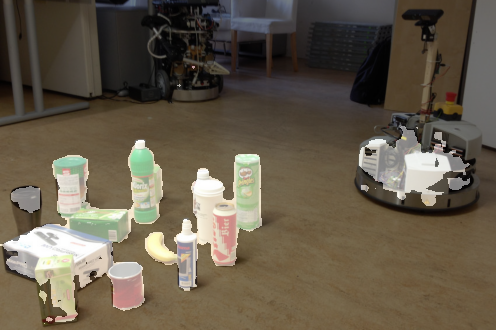}
    \includegraphics[height=2.6cm, trim={2cm 0 1cm 0},clip]{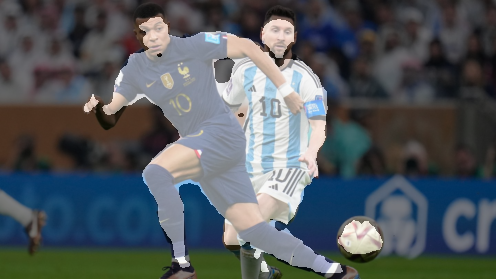}
    \includegraphics[height=2.6cm]{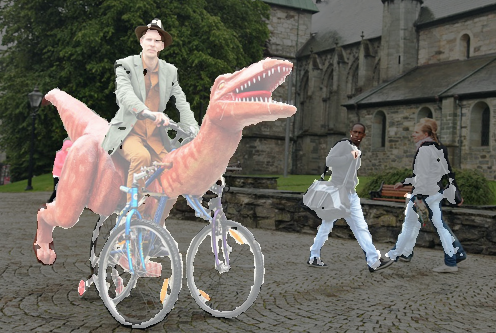}
    \\
\caption{\textbf{Visualization of \ours results on images taken from the Internet}. Objects out of the domain of ImageNet~\cite{jia2009imagenet} and DUT-TR \cite{wang2017} (datasets used for training the backbone and our segmentation head), of different scales, and of different shapes are correctly localized.}
\label{fig:open}
\end{figure*}

\begin{figure*}[!htp]
    \centering
     \small
    \resizebox{\textwidth}{!}{     
     \begin{tabular}{ccccc}
     \centering
         \includegraphics[width=3.12cm]{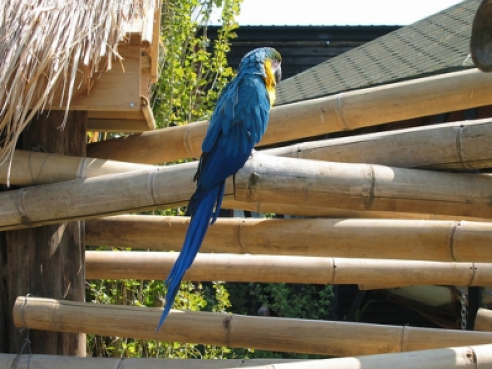} & 
         \includegraphics[width=3.12cm]{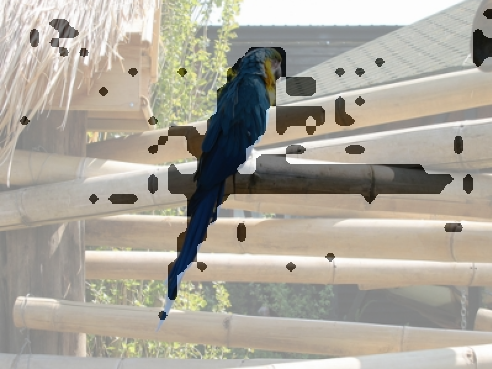} &  
         \includegraphics[width=3.12cm]{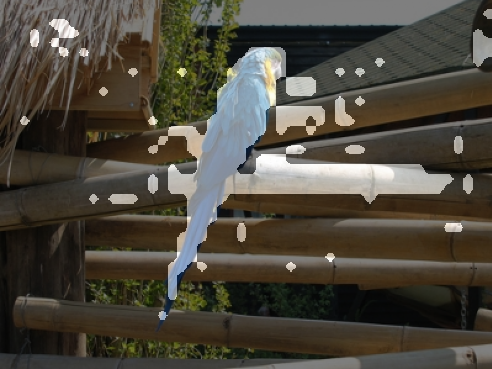} &
         \includegraphics[width=3.12cm]{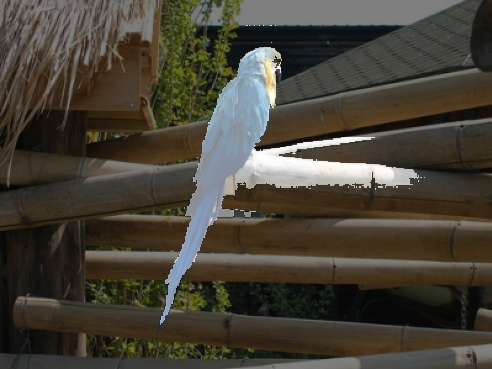} &
         \includegraphics[width=3.12cm]{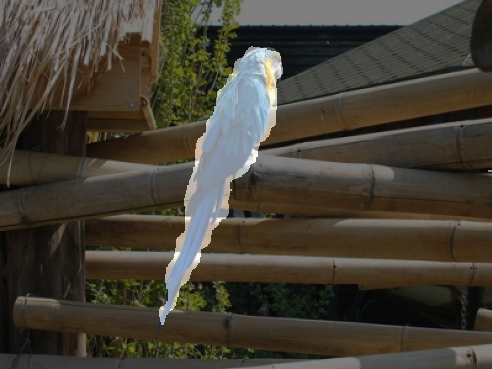} \\
         
         \includegraphics[width=3.12cm]{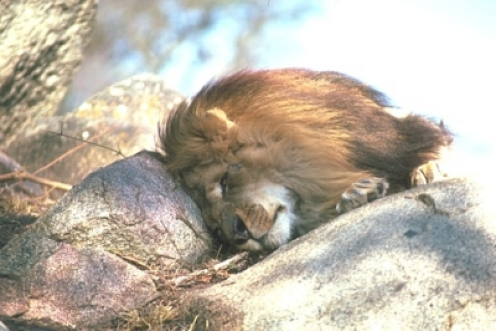} & 
         \includegraphics[width=3.12cm]{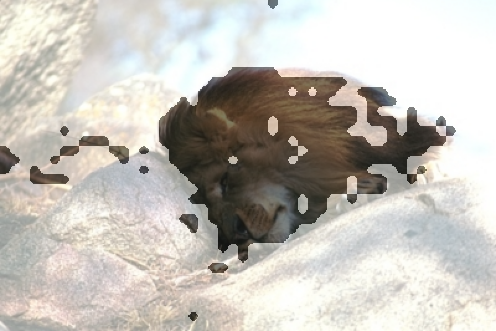} & 
         \includegraphics[width=3.12cm]{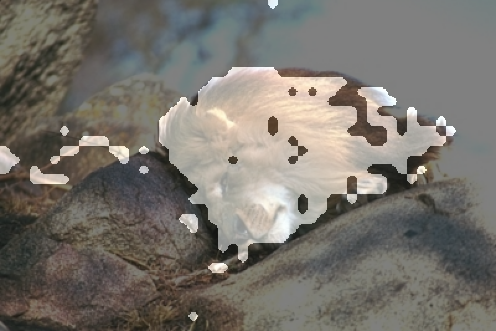} & 
         \includegraphics[width=3.12cm]{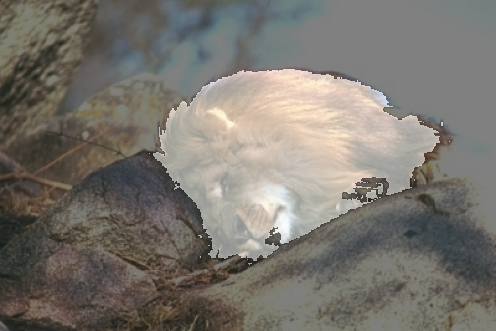} &
         \includegraphics[width=3.12cm]{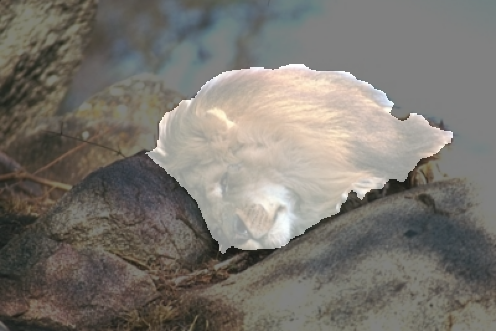} \\ 

         \includegraphics[width=3.12cm]{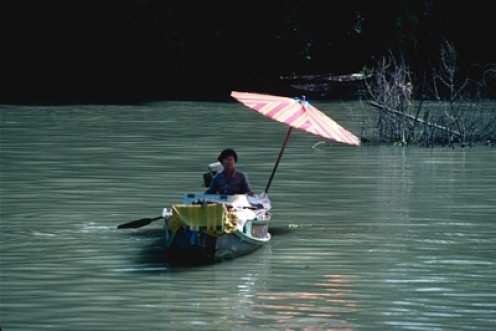} & 
         \includegraphics[width=3.12cm]{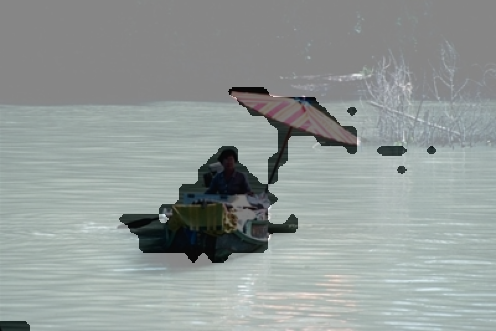} & 
         \includegraphics[width=3.12cm]{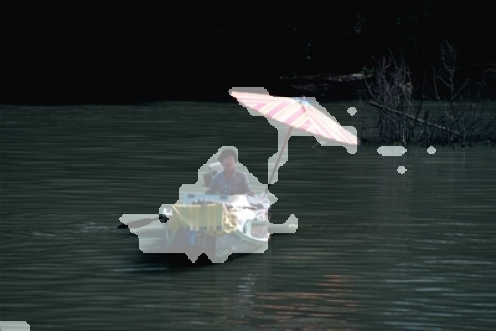} & 
         \includegraphics[width=3.12cm]{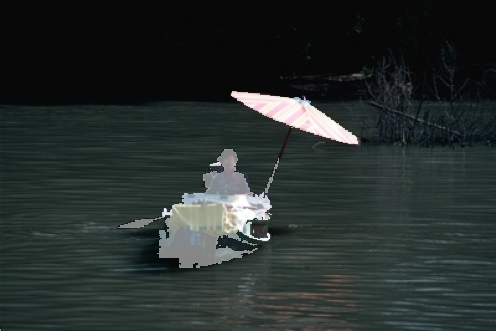} &
         \includegraphics[width=3.12cm]{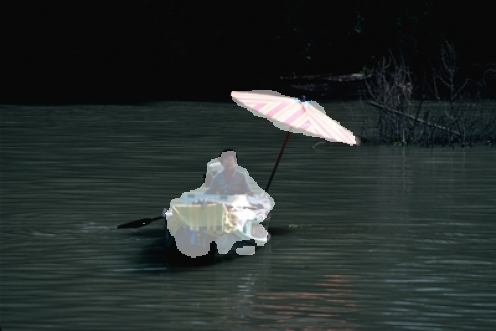} \\
         
        (a) Input image & (b) Coarse background & (c) Coarse foreground & 
        (d) Refined $\bs(\mathbf{M}^f)$ & (e) Predicted $\mathbf{M}^s$ \\
        & mask $\mathbf{M}^b$ & mask $\mathbf{M}^f$ & & (\ours)
    \end{tabular}
    }
    \caption{\textbf{More visualizations of masks generated on images from ECSSD \cite{shi2016ecssd} at different stages of our method}. We show (a) the input image, (b) the mask $\mathbf{M}^b$ extracted using our background discovery step, (c) its inverse $\mathbf{M}^f$ used as foreground mask to train our segmenter head, (d) the version refined using a bilateral solver $\zeta(\mathbf{M}^f)$, and (e) the final output of our trained segmentation head $\mathbf{M}^s$. 
    }
    \label{fig:sup-masks}
\end{figure*}

\subsection{Using masks from other methods}
We investigate here the performance of our method when considering different mask generators. In particular, we consider the well-known object discovery methods TokenCut \cite{wang2022tokencut} and LOST \cite{simeoni2021lost} with which we extract the masks $\mathbf{M}^f$ that are then refined in our training process (following \textcolor{red}{Sec. 3.2} of the main paper). We present the corresponding unsupervised object discovery results in \autoref{tab:coarse-masks}. They show that our method is agnostic to the mask generator but still performs slightly better with our foreground masks --- the complement of the background masks described in \textcolor{red}{Sec. 3.1}. It is also to be noted that our method is much faster than TokenCut because we do not need the computation of eigenvectors.

\section{Additional qualitative results}
We present in this section more visualizations of \ours results, first on more challenging images (\autoref{sec:random-im}) and at the different step of our process (\autoref{sec:masks-steps}). We then motivate the interest of reweighting the transformer heads (\autoref{sec:sup-heads}) via visual illustration. 
Following we show examples where the application of the bilateral solver impacts negatively the results (\autoref{sec:sup-bs}) and some more general failure cases of \ours (\autoref{sec:failures}).
We finally provide example of discovered objects as performed in the task of unsupervised object discovery (\autoref{sec:sup-uod}).

\subsection{Results on generic images from the Internet}
\label{sec:random-im}
We present in \autoref{fig:open} some results of \ours random images taken from the Internet. These results show the ability of \ours to discover multiple and diverse objects, both in terms of classes and scales. In particular, dinosaurs and spaceships are not depicted in ImageNet~\cite{jia2009imagenet} nor DUT-TR \cite{wang2017} and yet \ours can detect them, showing the ability to discover objects which ``are not background.'' Moreover, the selected images here are non-object centric and out-of-domain showing the capacity of \ours to go behond ImageNet-like images.

\begin{figure*}[!htp]
    \centering
     \small    
     \resizebox{\textwidth}{!}{     

     \begin{tabular}{ccccccc}
     \centering
         Input image & Head 0 & Head 1 & Head 2 & Head 3 & Head 4 & Head 5 \\
         \includegraphics[width=2.1cm]{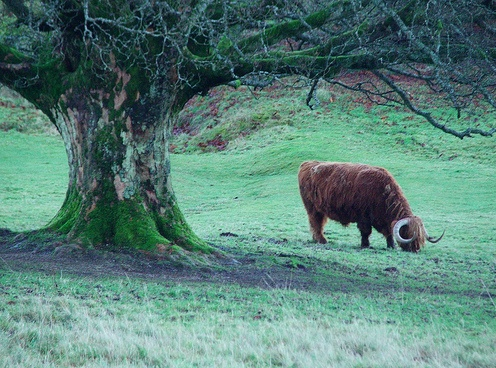} &
         \includegraphics[width=2.1cm]{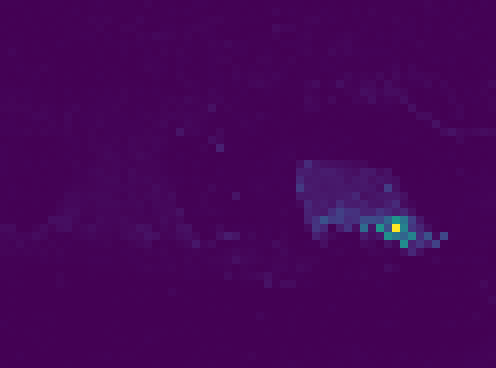} &
         \includegraphics[width=2.1cm]{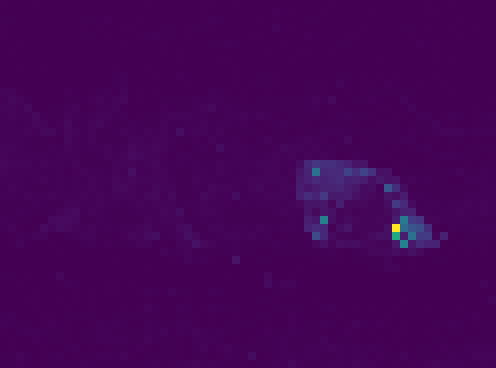} &
         \includegraphics[width=2.1cm]{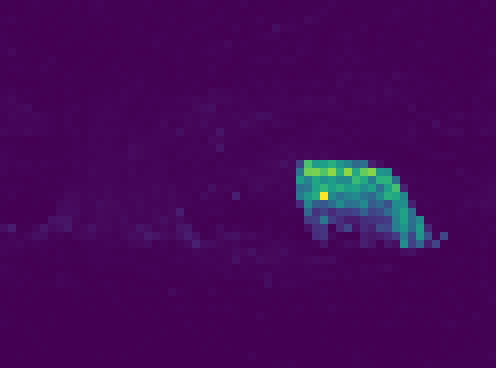} &
         \includegraphics[width=2.1cm]{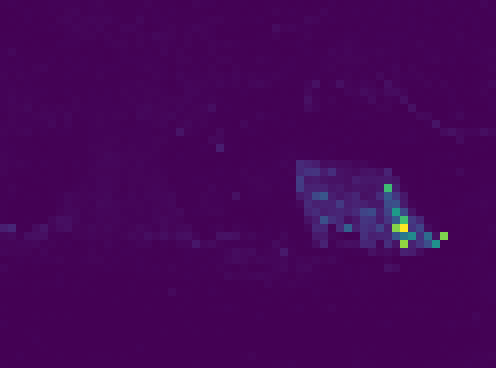} &
         \includegraphics[width=2.1cm]{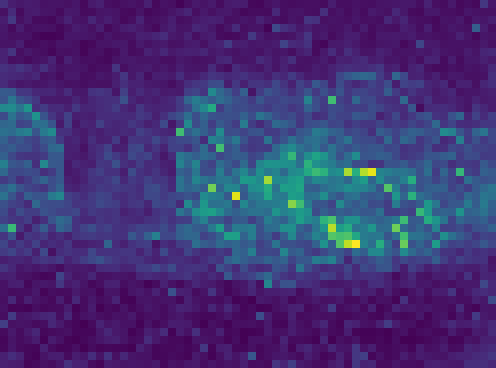} &
         \includegraphics[width=2.1cm]{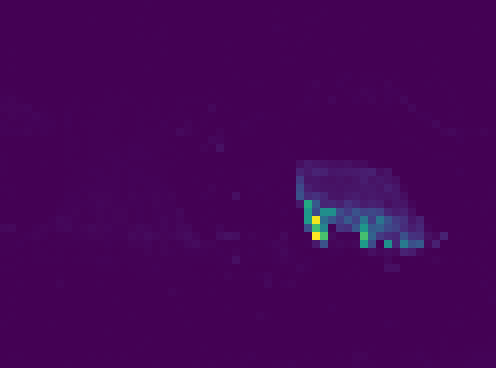} \\
         \includegraphics[width=2.1cm]{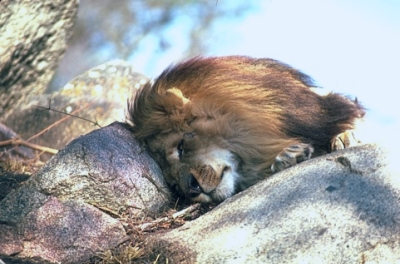} &
         \includegraphics[width=2.1cm]{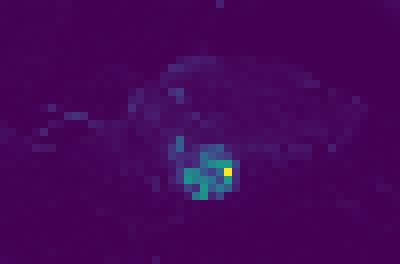} &
         \includegraphics[width=2.1cm]{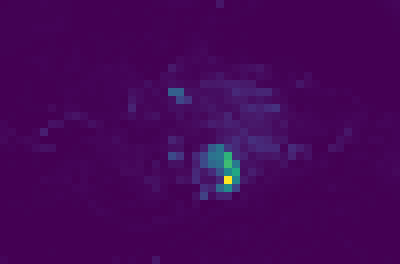} &
         \includegraphics[width=2.1cm]{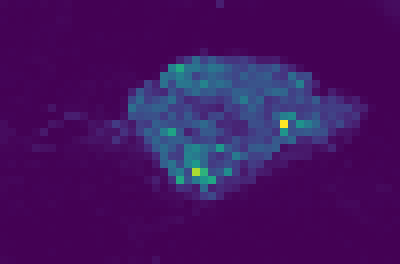} &
         \includegraphics[width=2.1cm]{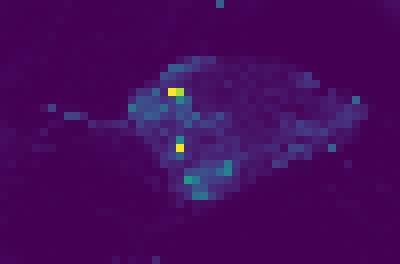} &
         \includegraphics[width=2.1cm]{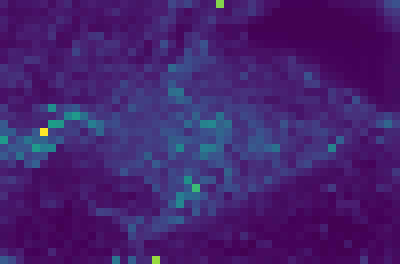} &
         \includegraphics[width=2.1cm]{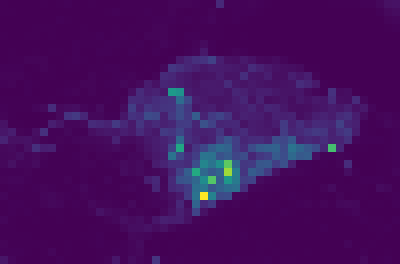}  
    \end{tabular}
    }
    \caption{\textbf{Visualization of self-attention maps} obtained with the six different heads in the last attention layer. Results are obtained with a ViT/S-8 trained using DINO~\cite{caron2021dino} applied on an image from VOC07 \cite{pascal-voc-2007} (first row) and ECSSD \cite{shi2016ecssd}.
    }
    \label{fig:sup-heads}
\end{figure*} 

\begin{figure}[!htp]
    \centering
     \small
     \begin{tabular}{cc}
     \centering
        Coarse mask before $\bs()$ & Coarse mask after $\bs()$ \\
        \includegraphics[width=3.5cm]{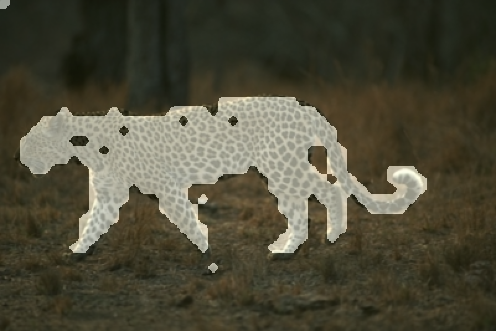} &
        \includegraphics[width=3.5cm]{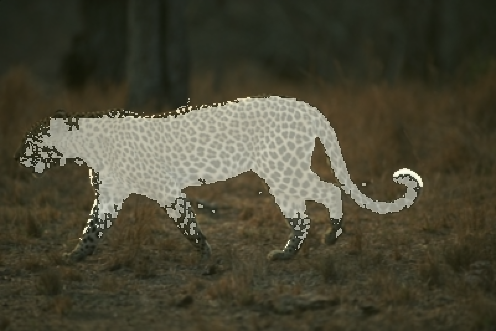} \\
        \includegraphics[width=3.5cm]{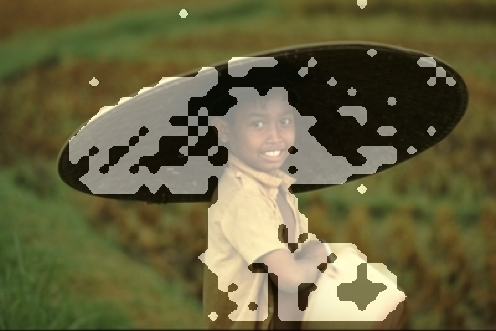} &
        \includegraphics[width=3.5cm]{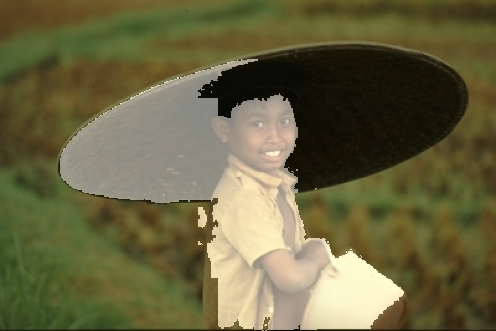} \\
        
         Pred. mask before $\bs()$ & Pred. mask after $\bs()$ \\
        \includegraphics[width=3.5cm]{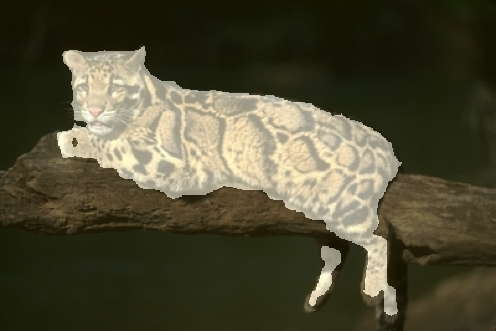} &
         \includegraphics[width=3.5cm]{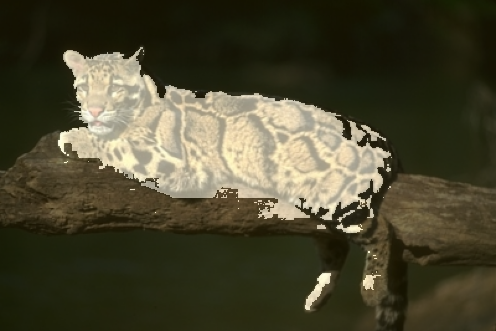} \\
         \includegraphics[width=3.5cm]{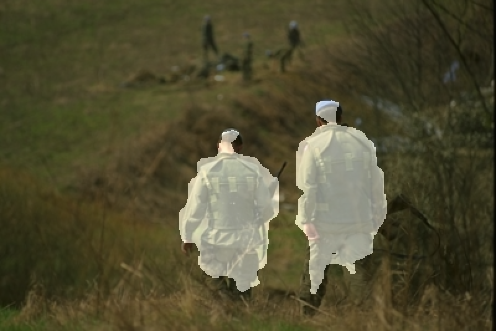} &
         \includegraphics[width=3.5cm]{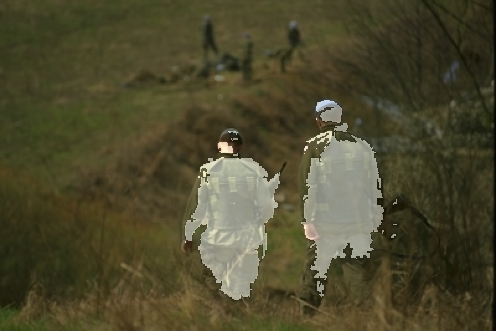} \\
    \end{tabular}
    \caption{\textbf{Visualization of the negative impact of the bilateral solver} on different ECSSD \cite{shi2016ecssd} images.}
    \label{fig:sup-bs}
\end{figure} 

\begin{figure}[!htp]
    \centering
     \small    
     \resizebox{\columnwidth}{!}{
    \begin{tabular}{ccc}
    Input image & \ours (ours) & SelfMask \cite{shin2022selfmask} \\
    \includegraphics[height=2.5cm,width=3.0cm]{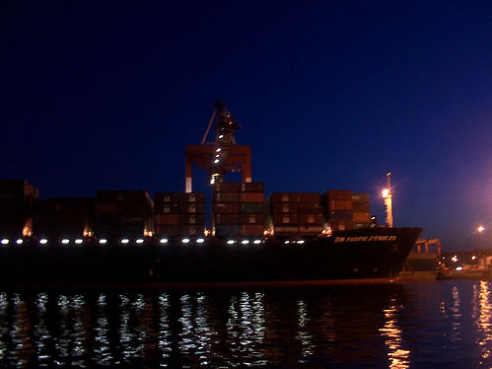} & \includegraphics[height=2.5cm,width=3.0cm]{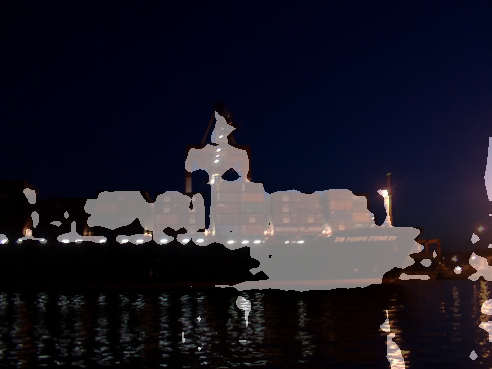} & 
    \includegraphics[height=2.5cm,width=3.0cm]{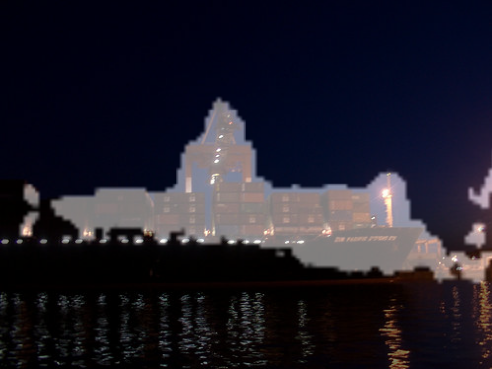} \\
    
    \includegraphics[height=2.5cm,width=3.0cm]{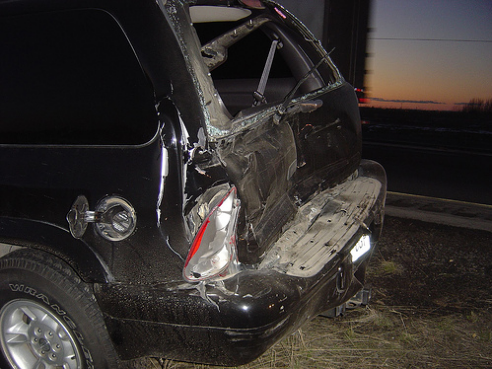} & 
    \includegraphics[height=2.5cm,width=3.0cm]{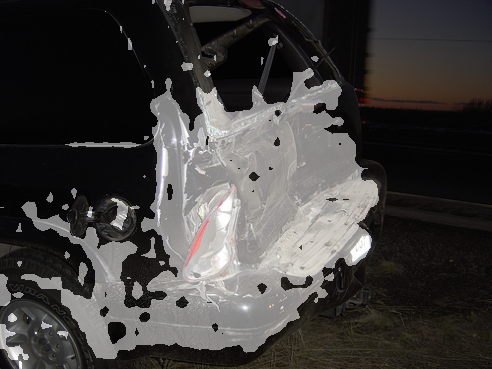} & 
    \includegraphics[height=2.5cm,width=3.0cm]{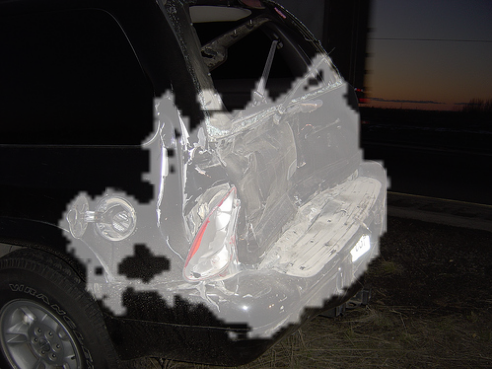}  \\
    
    \includegraphics[height=2.5cm,width=3.0cm]{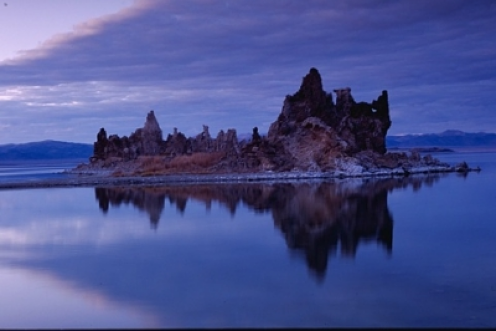} & 
    \includegraphics[height=2.5cm,width=3.0cm]{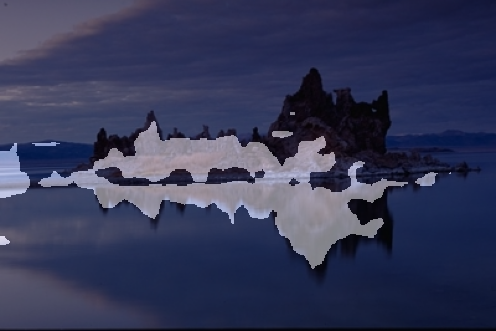} & 
    \includegraphics[height=2.5cm,width=3.0cm]{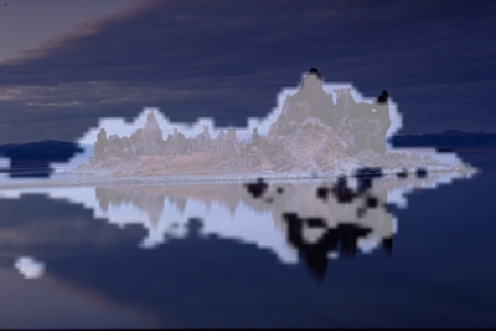} \\
    
    \includegraphics[height=2.5cm,width=3.0cm]{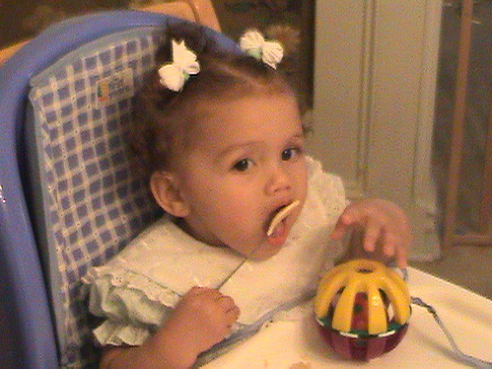} & 
    \includegraphics[height=2.5cm,width=3.0cm]{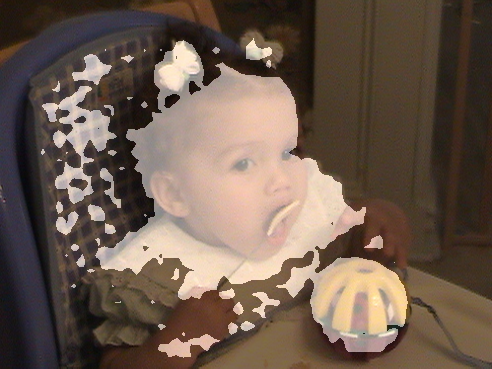} & 
    \includegraphics[height=2.5cm,width=3.0cm]{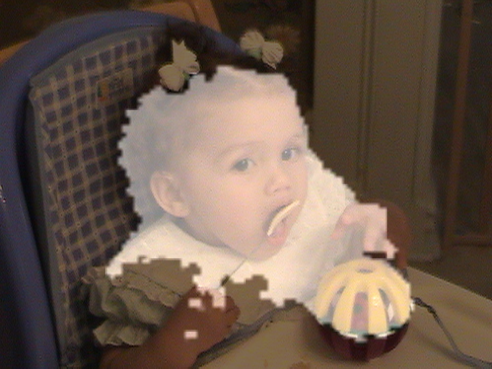} 

    \end{tabular}
    }
    \caption{\textbf{Visualization of failure cases for object localization} on images from ECSSD \cite{shi2016ecssd}, PASCAL VOC07 \& VOC12 \cite{pascal-voc-2007, pascal-voc-2012} and DUTS-TE \cite{wang2017} datasets along side results obtained with SelfMask method. No refinement step is applied.}
    \label{fig:failures_ours-new}
\end{figure}

\subsection{Visualization of masks at different steps}
\label{sec:masks-steps}
We provide in \autoref{fig:sup-masks} additional visualizations of the masks generated at different steps of our method. We can observe that each step brings an improvement over the previous one. The right-most column presents the final output of \ours without any refinement.

\subsection{Reweighting the attention heads}
\label{sec:sup-heads}
We provide in \autoref{fig:sup-heads} a visualization of the self-attention maps extracted from the last layer of our model. We show the self-attention obtained over the six heads; we can observe that the $4^{\rm th}$ head is noisy.
When looking for the background \emph{seed}, we are looking for the pixel with least attention. Our reweighting scheme helps in reducing the weight given to such noisy heads automatically and improves results, as shown in \textcolor{red}{Tab. 5} of the main paper.

\subsection{Potential negative effect of the bilateral solver}
\label{sec:sup-bs}
While the application of $\bs()$, the bilateral solver \cite{bilateralsolverbarron2016}, improves results in general (see \autoref{fig:sup-masks}), there are cases where $\bs()$ actually degrades the mask quality. We show examples of such cases in \autoref{fig:sup-bs} both on coarse masks (rows 1 and 2) and on the final outputs (rows 3 and 4). We can observe that the function amplifies the under-segmentation, \eg, on the hat and the leopard head and legs (row 1 and 2). Moreover, long and thin segments can disappear, \eg, human and animal legs or arms (row 3). Correcting this behaviour would help improving our training and is left for future work.

\subsection{Examples of failures cases}
\label{sec:failures}
We show some failure cases of \ours in \autoref{fig:failures_ours-new}. For these cases, we also present the results obtained with one of the best competitor:  SelfMask \cite{shin2022selfmask}.
We observe that night or dark scenes are challenging (first {two rows}). Our method tends to under-segment objects but SelfMask has also difficulties in segmenting correctly the main objects in these situation. \ours, just like SelfMask, is also not robust to reflection on water (third row). Finally, we observe that both methods fail to segment the hair in the fourth columns.

\subsection{Unsupervised object discovery results}
\label{sec:sup-uod}
\begin{figure*}[!htp]
    \small
    \centering
    \begin{tabular}{ccccc}
    (a) Ground truth & (b) \ours{} (ours) & (c) TokenCut \cite{wang2022tokencut} & (d) SelfMask \cite{shin2022selfmask} & (e) FreeSOLO \cite{wang2022freesolo} \\
    \includegraphics[width=3cm]{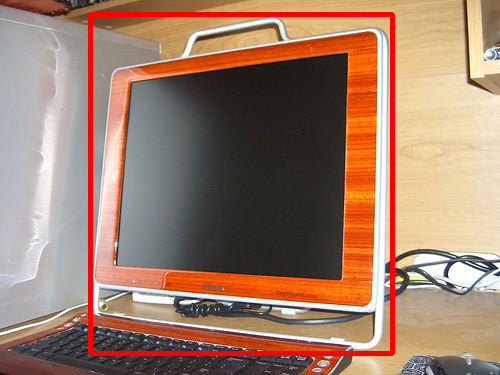} & \includegraphics[width=3cm]{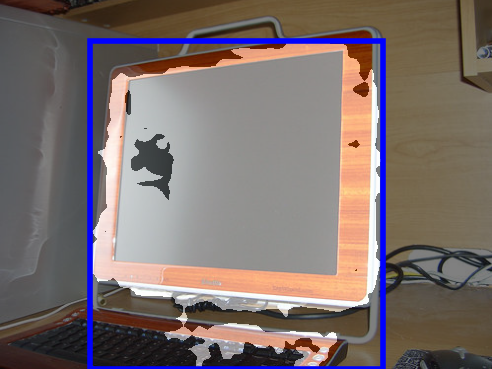} & \includegraphics[width=3cm]{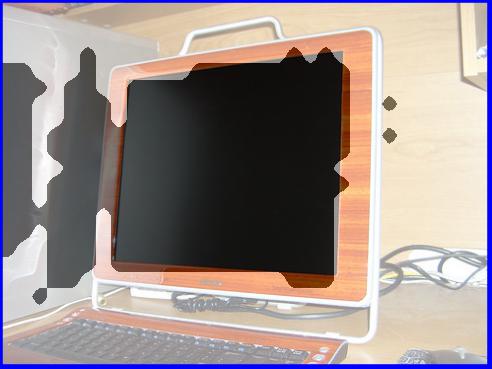} & \includegraphics[width=3cm]{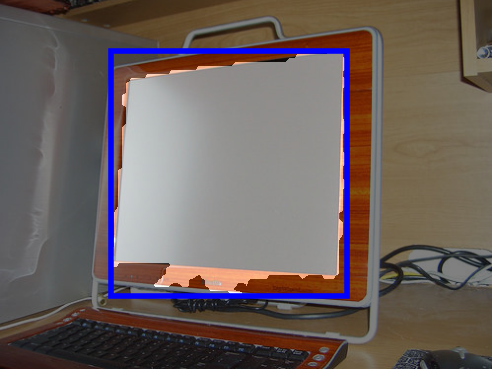} & \includegraphics[width=3cm]{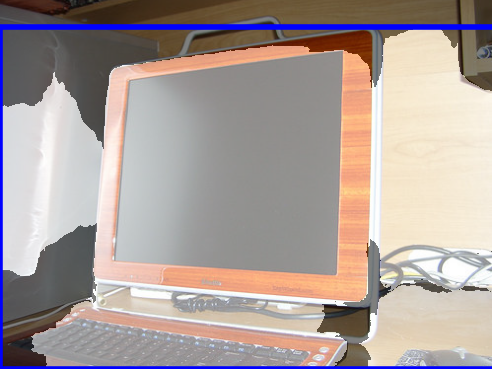}\\
    \includegraphics[width=3cm]{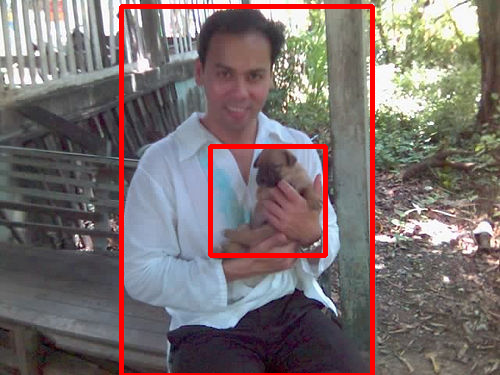} &  \includegraphics[width=3cm]{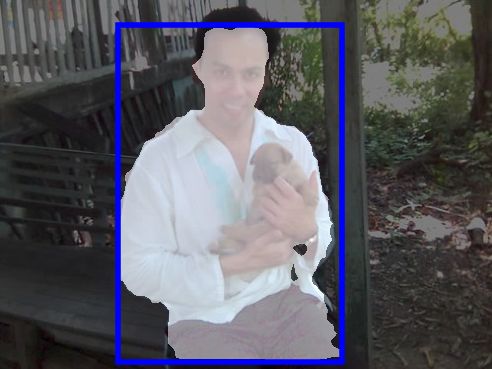} & \includegraphics[width=3cm]{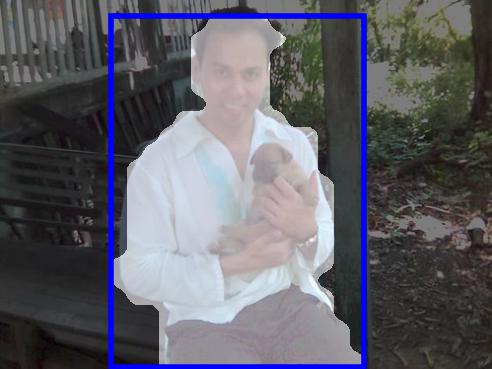} & \includegraphics[width=3cm]{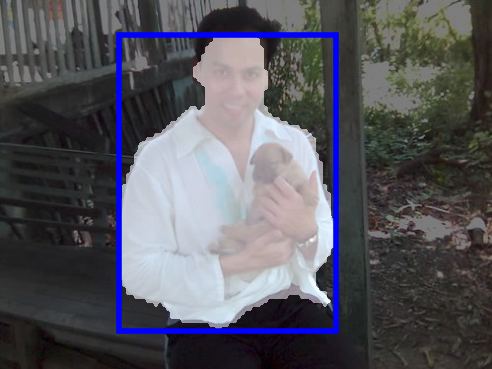} & \includegraphics[width=3cm]{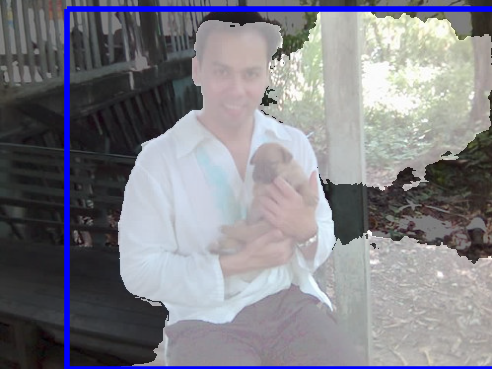}\\
    \includegraphics[width=3cm]{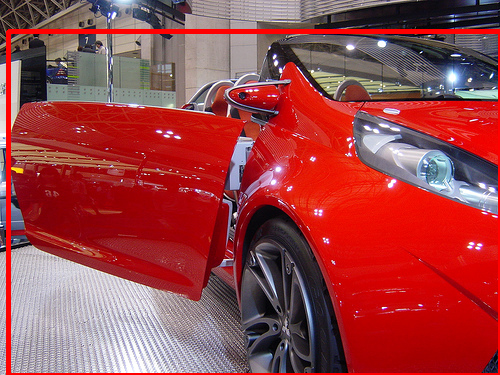} &  \includegraphics[width=3cm]{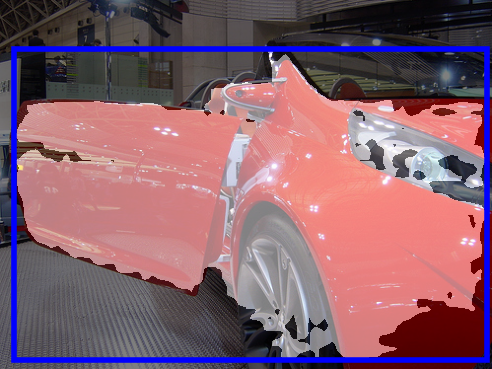} & \includegraphics[width=3cm]{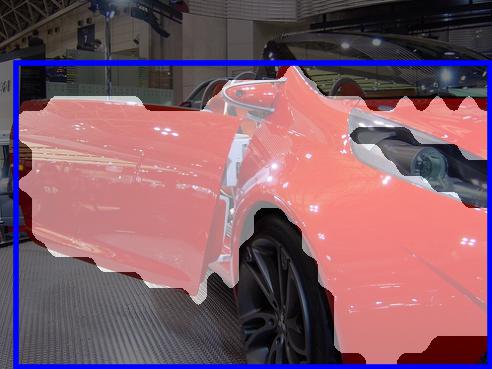} & \includegraphics[width=3cm]{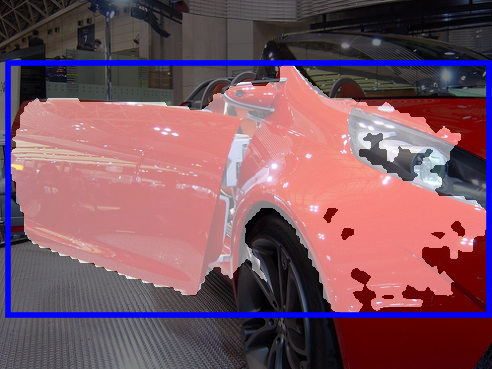} & \includegraphics[width=3cm]{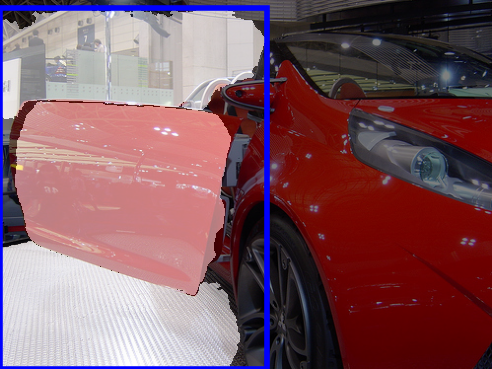}\\
    \includegraphics[width=3cm]{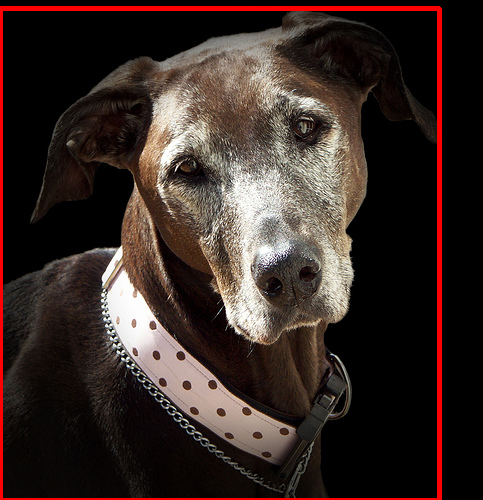} &  \includegraphics[width=3cm]{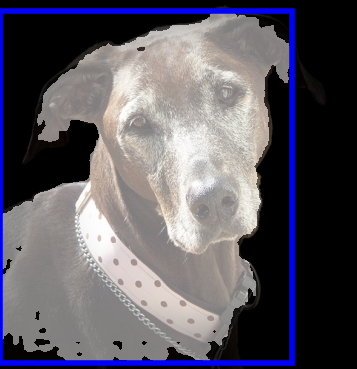} & \includegraphics[width=3cm]{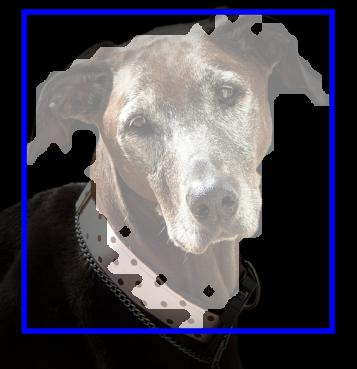} & \includegraphics[width=3cm]{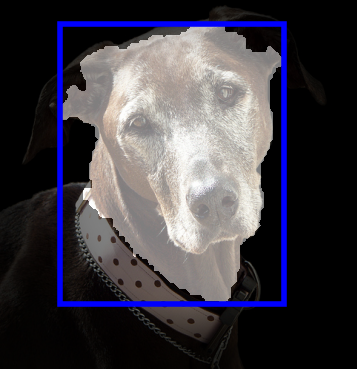} & \includegraphics[width=3cm]{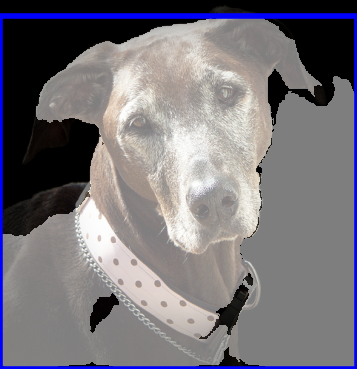}\\
    \includegraphics[width=3cm]{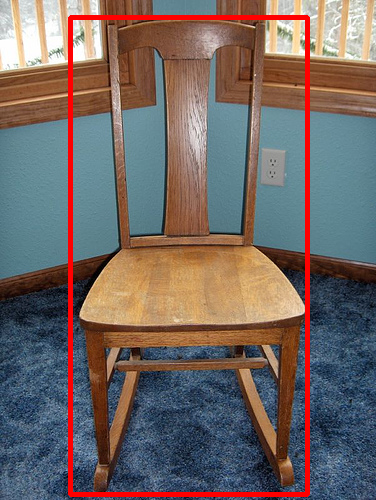} &  \includegraphics[width=3cm]{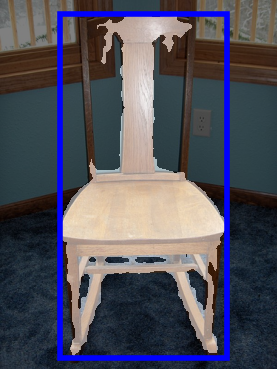} & \includegraphics[width=3cm]{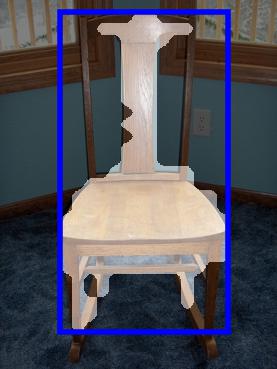} & \includegraphics[width=3cm]{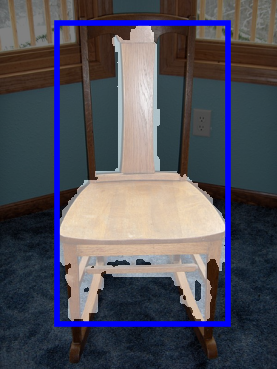} & \includegraphics[width=3cm]{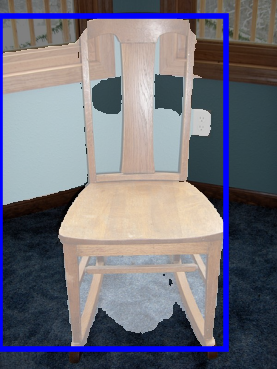}\\
    \includegraphics[width=3cm]{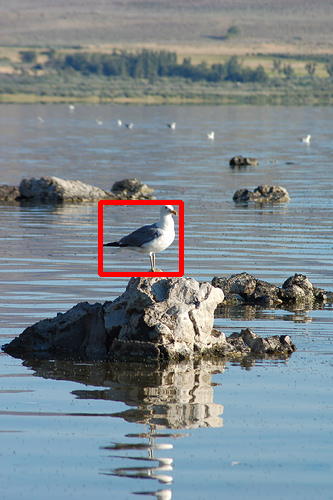} &  \includegraphics[width=3cm]{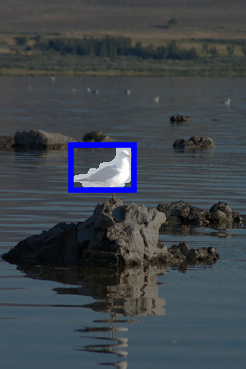} & \includegraphics[width=3cm]{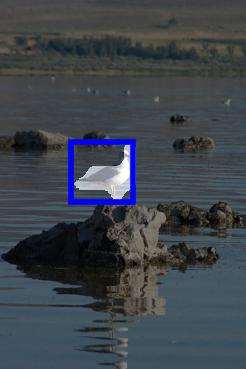} & \includegraphics[width=3cm]{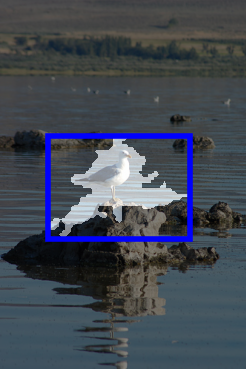} & \includegraphics[width=3cm]{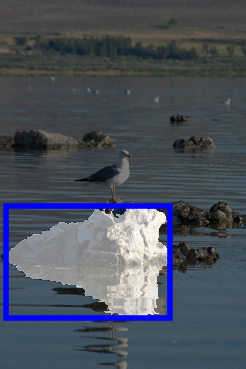}
    \end{tabular}
    \caption{\textbf{Qualitative results} for the task of \emph{unsupervised single object discovery} on PASCAL VOC12 dataset \cite{pascal-voc-2012}. We show here masks and boxes extracted as defined in \textcolor{red}{Sec. 4.1}. In particular, \ours is in the \emph{single} setup (\ours{} -- single). No refinement step is applied on the masks. }
    \label{fig:sod_viz}
\end{figure*}

We present in \autoref{fig:sod_viz}, qualitative results for the unsupervised single object discovery task (no refinement is applied to the masks). We draw the extracted bounding box on top of the corresponding predicted mask.
The conclusions here are similar to those discussed in the main paper. Overall our method segments the objects of interest better and provides cleaner boundaries.

\end{document}